\documentclass[runningheads]{llncs}

\usepackage{eccv}

\usepackage{eccvabbrv}

\usepackage{graphicx}
\usepackage{amsmath}
\usepackage{amssymb}
\usepackage{booktabs}

\usepackage{makecell}
\usepackage{colortbl}
\usepackage{array} 

\usepackage[accsupp]{axessibility}  % Improves PDF readability for those with disabilities.

\usepackage{colortbl}  %彩色表格需要加载的宏包
\usepackage{xcolor}

\usepackage{marvosym} 
\usepackage{multirow}

\usepackage{ulem}
\usepackage{amsmath}
\usepackage{xcolor}
\usepackage{caption}

\newcommand{\Paragraph}[1]{\noindent\textbf{#1}}
\def\red#1{\textbf{\color{red}  #1}} %in Table

\renewcommand{\ie}{\textit{i.e.}~}
\renewcommand{\eg}{\textit{e.g.}~}

\usepackage{hyperref}

\usepackage{orcidlink}

\begin{document}

\title{CONDA: Condensed Deep Association Learning for Co-Salient Object Detection} 

\titlerunning{CONDA}

\author{Long Li\inst{1}\orcidlink{0000-0002-1939-5941},
Nian Liu\inst{3,*}\orcidlink{0000-0002-0825-6081},
Dingwen Zhang\inst{1}\orcidlink{0000-0001-8369-8886},
Zhongyu Li\inst{4},
Salman Khan\inst{3,5}\orcidlink{0000-0002-9502-1749},
Rao Anwer\inst{3}\orcidlink{0000-0002-9041-2214}, 
Hisham Cholakkal\inst{3}\orcidlink{0000-0002-8230-9065},
Junwei Han\inst{1,2,*}\orcidlink{0000-0001-5545-7217}, and \\ Fahad Shahbaz Khan\inst{3,6}\orcidlink{0000-0002-4263-3143}
}

% \thanks{*Corresponding authors: \{junweihan2010, liunian228\}@gmail.com}
\footnotetext{${}^*$ Corresponding authors: \{liunian228, junweihan2010\}@gmail.com}

\authorrunning{L Li et al.}

\institute{Northwestern Polytechnical University\and
Institute of Artificial Intelligence, Hefei Comprehensive
National Science Center\and
Mohamed bin Zayed University of Artificial Intelligence
\and
Xi'an Jiaotong University \and
Australian National University
\and
CVL, Linköping University
}

\maketitle

\begin{abstract}
Inter-image association modeling is crucial for co-salient object detection. Despite satisfactory performance, previous methods still have limitations on sufficient inter-image association modeling. Because most of them focus on image feature optimization under the guidance of heuristically calculated raw inter-image associations. They directly rely on
raw associations which are not reliable in complex scenarios, and their image feature optimization approach is not explicit for inter-image association modeling. To alleviate these limitations, this paper proposes a deep association learning strategy that deploys deep networks on raw associations to explicitly transform them into deep association features. Specifically, we first create hyperassociations to collect dense pixel-pair-wise raw associations and then deploys deep aggregation networks on them. We design a progressive association generation module for this purpose with additional enhancement of the hyperassociation calculation. More importantly, we propose a correspondence-induced association condensation module that introduces a pretext task, \ie semantic correspondence estimation, to condense the hyperassociations for computational burden reduction and noise elimination. We also design an object-aware cycle consistency loss for high-quality correspondence estimations. Experimental results in three benchmark datasets demonstrate the remarkable effectiveness of our proposed method with various training settings. The code is available at: \href{https://github.com/dragonlee258079/CONDA}{https://github.com/dragonlee258079/CONDA}.
\end{abstract}

\section{Introduction}
% \vspace{-2mm}
\label{sec:intro}

\begin{figure}[t]
\centering
\includegraphics[width=1.0\linewidth]{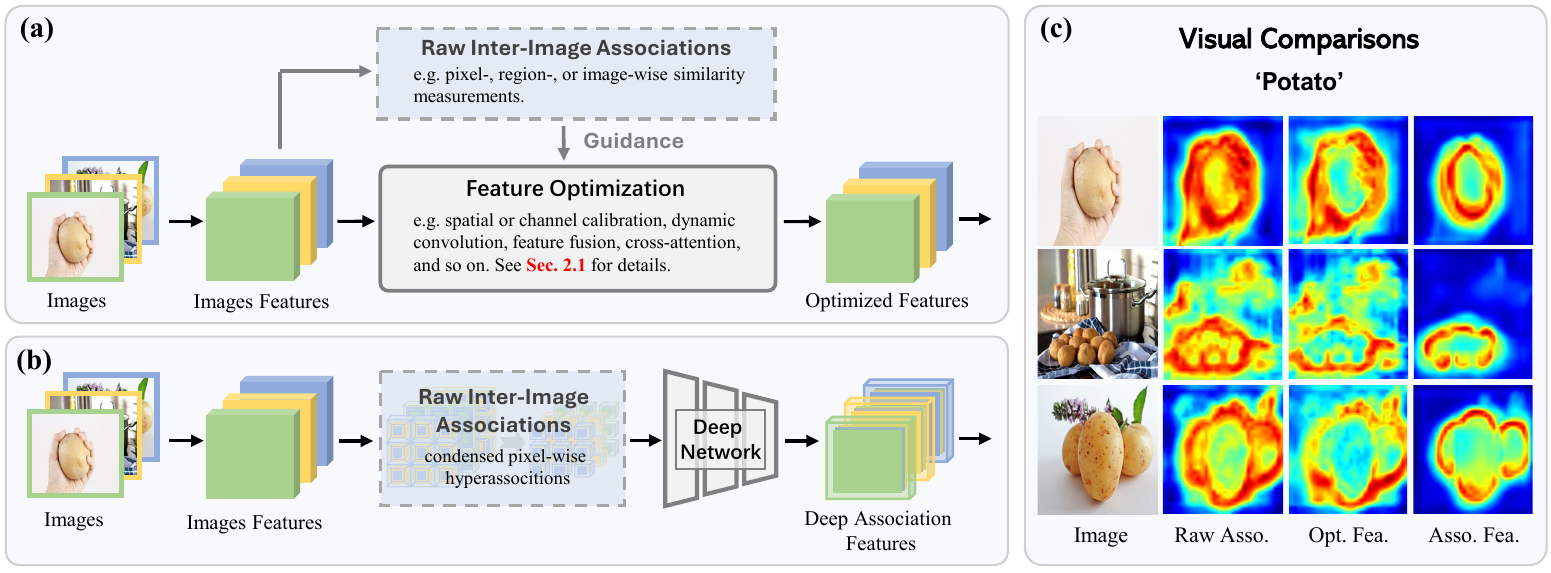}
\vspace{-7mm}
\caption{\textbf{Difference of raw-association-based image feature optimization strategy (a) and our proposed deep association learning strategy (b).} Our deep association learning deploys deep learning networks on raw associations to achieve deep association features. We also present visual samples of our calculated raw associations (Raw Asso.), optimized image feature (Opt. Fea.), and our generated deep association features (Asso. Fea.) in (c).}
\label{introduction}
\vspace{-6mm}
\end{figure}

Co-Salient Object Detection (CoSOD) aims to segment salient objects that appear commonly across a group of related images. Compared to traditional Salient Object Detection (SOD) \cite{liu2016dhsnet,liu2018picanet,zhao2019egnet,liu2019simple,liu2021visual,zhao2021weakly,zhuge2022salient,liu2024vst++,luo2024vscode}, CoSOD is a more challenging task because it requires 
sufficient inter-image association modeling  \cite{fu2013cluster}. 

Recently, many advanced works \cite{zhang2020gicd,jin2020icnet,fan2021GCoNet,zhang2021summarize,su2023unified,yu2022democracy,li2023discriminative,zheng2023gconet+,wu2023co} have emerged and achieved impressive performance.
These methods
first use related image features to acquire raw inter-image associations (also called consensus representations), and then leverage them as guidance to optimize each image feature, as shown in Figure~\ref{introduction} (a). This approach enables the final image feature to implicitly capture inter-image cues, thereby achieving the purpose of inter-image association modeling.
However, we find this \textit{raw-association-based feature optimization} strategy still has two limitations: 
i) they directly rely on raw associations, which are acquired in heuristic manners, such as pixel-wise \cite{fan2021GCoNet,yu2022democracy,su2023unified,zheng2023gconet+}, region-wise \cite{zhang2021summarize,li2023discriminative}, or image-wise \cite{jin2020icnet} similarity measurements. 
Although high-quality raw associations can be derived from high-level semantic information in image features, 
their revelation of common saliency regions still relies on similarity measures, which are unreliable when encountering complex scenarios, such as significant differences between co-salient objects or high foreground-background similarity.
ii) the primary focus of building their deep models is on optimizing image features. Compared to directly modeling association relationship, image feature optimization is not an explicit approach for inter-image association modeling and will increase the learning difficulty.

To alleviate these limitations, we propose a \textit{deep association learning} strategy for CoSOD, as shown in Figure~\ref{introduction} (b). Instead of directly using raw associations to optimize image features, we deploy deep networks on raw associations to learn deep association features. This is a more explicit strategy for inter-image association modeling. Moreover, our deep association features can capture high-level inter-image association knowledge, making them more robust in complex scenarios than raw associations, as shown in Figure~\ref{introduction} (c).  
Technically, we start by collecting all pixel-wise raw associations across the entire image group as hyperassociations. Then, we propose a Progressive Association Generation (PAG) module to transform hyperassociations into deep association features. PAG progressively generates association features on varying scales, allowing us to use previous association features to enhance the hyperassociation calculation at the next scale, thereby improving the association quality from the very beginning.

Although deep association learning strategy allows more sufficient inter-image association modeling,
it significantly increases the computational burden and reduces the practicality of this approach. Additionally, this study finds that it is not necessary to utilize all pixel associations to generate deep association features. In fact, there are even some noisy pixels that negatively impact the quality of the deep association features. Therefore, we propose a method based on Correspondence-induced Association Condensation (CAC) to condense the original full-pixel hyperassociations. This not only alleviates the computational burden but also further enhances the quality of deep association features. 

Specifically, CAC performs the condensation operation by selectively associating pixels that have semantic correspondence in other images, as well as their surrounding contextual pixels, thereby creating lightweight yet more accurate hyperassociations. 
Here, we introduce a pretext task, \ie semantic correspondence estimation, into the CoSOD, not only improving the model performance but also delving deeper into the essence of CoSOD. Co-salient objects inherently possess \textit{object-level} semantic correspondence. However, in this paper, we aim to further explore the finer \textit{pixel-level} correspondence. Although highly accurate correspondence estimation remains a challenge, we believe it will pave a new way for CoSOD research. We also provide an object-aware cycle consistency (OCC) loss to aid in learning correspondences within co-salient pixels.

In summary, the contributions of this paper are as follows:
\vspace{-1.5mm}
\begin{itemize}
\item For the first time, we introduce a \textit{deep association learning} approach for CoSOD, applying deep networks to transform raw associations into deep association features for sufficient inter-image association modeling. Specifically, we develop a \textbf{CON}densed \textbf{D}eep \textbf{A}ssociation (CONDA) learning model. 
% \vspace{-1mm}
\item We propose a PAG module to progressively generate deep association features. It enhances image features with previous association features to improve hyperassociation calculation.  
% \vspace{-1mm}
\item We introduce semantic correspondence estimation into the CoSOD task to condense the original hyperassociation for alleviating the computational burden and further improving the performance. 
We also propose an OCC loss for effective correspondence estimation.
% \vspace{-1mm}
\item Experimental results demonstrate that our model achieves significantly improved state-of-the-art performance on three benchmark datasets across different training settings.
\end{itemize}

\section{Related Work}
% \vspace{-2mm}
\subsection{Co-Salient Object Detection}
\label{related_work_cosod}

Recently, there has been a surge of excellent methods for CoSOD \cite{zhang2018review, han2017unified, zhang2016co, zhang2015cosaliency, zhang2020gicd,jin2020icnet,fan2021GCoNet,zhang2021summarize,su2023unified,yu2022democracy,li2023discriminative,zheng2023gconet+,wu2023co}. These methods
initially acquire raw inter-image associations with related image features and then utilize them to optimize each image feature. Most methods generate pixel-wise \cite{fan2021GCoNet,yu2022democracy,su2023unified,zheng2023gconet+}, region-wise \cite{zhang2021summarize,li2023discriminative}, and image-wise \cite{jin2020icnet} raw associations through similarity-based manners, \eg inner product calculations between image features. 
Even for transformer-based methods \cite{su2023unified,li2023discriminative}, they also rely on inner product calculation to produce attention maps \cite{vaswani2017attention} as raw associations.
The image feature optimization manners include spatial or channel calibration \cite{jin2020icnet, fan2021GCoNet, zheng2023gconet+}, dynamic convolution \cite{zhang2021summarize, yu2022democracy}, feature fusion \cite{yu2022democracy, li2023discriminative}, and cross-attention \cite{su2023unified, li2023discriminative}, etc. However, they lack the learning of high-level association knowledge and heavily focus on optimized image features. Unlike them, this paper proposes a new research direction that deploys deep networks on associations to achieve deep association features for CoSOD.

% \vspace{-3mm}
\subsection{Inter-image Relation Modeling}
% In addition to the CoSOD task, there are several other tasks that require considering the relationships between multiple images,
Apart from CoSOD, other tasks necessitating the consideration of inter-image relationships,
such as few-shot segmentation \cite{min2021hypercorrelation, liu2022intermediate, moon2022hm, xiong2022doubly}, stereo matching \cite{chen2023costformer, xu2023iterative}, video semantic segmentation \cite{sun2022mining},
etc.
% These tasks have made remarkable progress in recent years, thanks to their ability to effectively model the inter-image relations. 
These tasks have made significant advancements recently by effectively modeling inter-image relations.
% Most of them \cite{min2021hypercorrelation, xiong2022doubly, chen2023costformer, xu2023iterative, cho2021cats, hong2022neural} first maintain cost volumes to capture dense inter-image pixel-wise similarities and then deploy various modules to transform this hyper volume to their task-specific features. Our work differs from previous methods in three points.
Most of these methods \cite{min2021hypercorrelation, xiong2022doubly, chen2023costformer, xu2023iterative, cho2021cats, hong2022neural} initially create cost volumes to capture dense inter-image pixel-wise similarities and subsequently use various modules to convert these hyper volumes into task-specific features. 
Our approach distinguishes itself from prior methods in three aspects.
Firstly, most of them create 4D cost volumes between two images while we create 6D hyperassociations between all related images. Secondly, they calculate hyper volumes based on original image features, whereas We propose PAG to progressively enhance image features for better hyperassociation calculation. Last and most importantly, they rely on the full-pixel cost volume. We consider the condensation of hyperassociations using semantic correspondences to eliminate noisy pixel associations. 

% \vspace{-3mm}
\subsection{Semantic Correspondence Estimation}
Semantic correspondence estimation \cite{liu2020semantic} aims to establish reliable pixel correspondences between different instances of the same object category. 
Most works performed this task using fully supervised training \cite{liu2020semantic, kim2022transformatcher}.
Some recent works have utilized unsupervised learning with photometric, forward-backward consistency, and warp-supervision losses \cite{shen2020ransac, truong2021warp, zhao2021multi}. However, they implement these losses on the entire image, where background pixels may affect the performance. In this paper, we introduce this task to condense hyperassociations for Co-SOD and tailor the cycle consistency loss by only applying it to co-salient pixels, hence effectively avoiding the influence of background and extraneous objects. 

\begin{figure*}[t]
\centering
\includegraphics[width=1\linewidth]{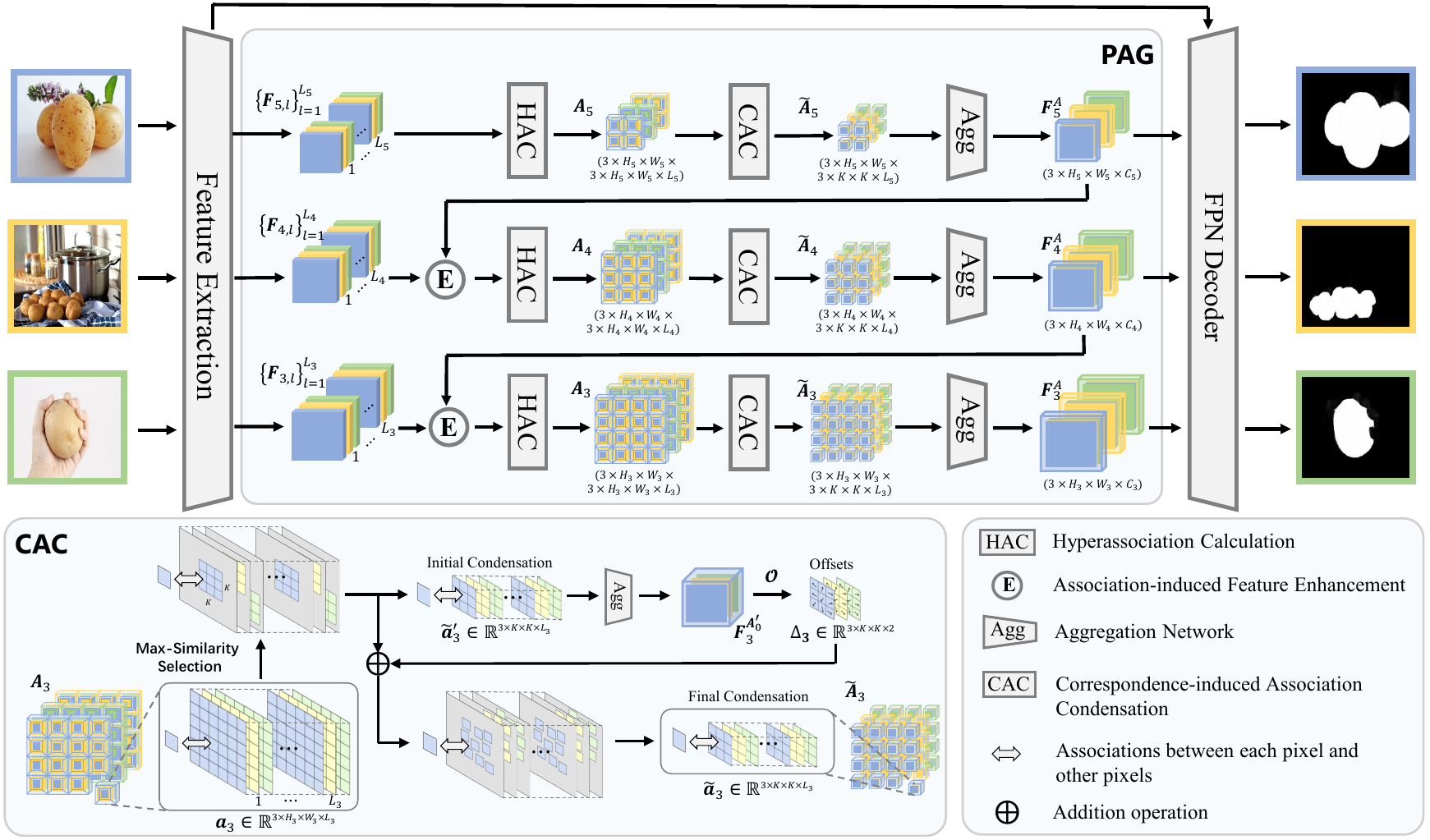}
\vspace{-6mm}
\caption{\textbf{Overall flowchart of our CONDA model.} Specifically, CONDA first utilizes the image features to calculate hyperassociations. Then, the full-pixel hyperassociations are condensed by CAC and fed into the aggregation networks to achieve deep association features. These features are then used in the FPN decoder process for the final prediction. To be concise, only three related images are shown.
}

\label{framework}
\vspace{-6mm}
\end{figure*}

% \vspace{-2mm}
\section{Proposed Method}
% \vspace{-1mm}
% \subsection{Overview}
% \vspace{-1mm}
As shown in Figure~\ref{framework}, CONDA integrates the deep association learning process into an FPN framework. Specifically, given a group of related images $\{\boldsymbol{I}_i\}_{i=1}^{N}$, we first input them into a VGG-16 \cite{simonyan2014very} backbone to collect its intermediate features for PAG and FPN decoding. In detail, we collect all features of the last three stages for PAG and the last feature of each stage for the FPN decoder as follows:
\begin{equation}
\begin{gathered}
\mathbb{F}^{P} = \big\{\boldsymbol{F}_{s,l} \,|\, s \in \{3, \cdots, 5\}, l \in \{1, \cdots, L_s\}\big\}, \\
\mathbb{F}^{D} = \big\{\boldsymbol{F}_{s,L_s} \,|\, s \in \{1, \cdots 5\}\big\},
\end{gathered}
\end{equation}
where $\mathbb{F}^{P}$ and $\mathbb{F}^{D}$ are the feature collections for PAG and FPN decoder, respectively. $\boldsymbol{F}_{s,l} \in \mathbb{R}^{N \times H_s \times W_s \times C_s}$ represents the VGG feature of the $l$-th layer in the $s$-th stage. There are in total $L_s$ layers in the $s$-th stage. $H_s$, $W_s$ and $C_s$ represent the height, width, and channel of the $s$-th stage, respectively.

Then, we input $\mathbb{F}^P$ into PAG to calculate hyperassociations and genarate deep association features $\{\boldsymbol{F}_s^A \in \mathbb{R}^{N\times H_s \times W_s \times C_s}\}_{s=3}^5$. Finally, these association features will be fused with $\mathbb{F}^D$ for the FPN decoder process, formulated as:
\begin{equation}
\begin{gathered}
\boldsymbol{F}_{s,L_s} = \boldsymbol{F}_{s,L_s} + \phi(\boldsymbol{F}_s^A), \,\,\,\,\, s=3,\cdots,5,\\
\boldsymbol{F} = \operatorname{Decoder}(\{\boldsymbol{F}_{s,L_s}\}_{s=1}^5),
\end{gathered}
\end{equation}
where $\phi$ is a convolution layer. $\boldsymbol{F} \in \mathbb{R}^{N \times H \times W \times C}$ is the final feature for the final co-saliency prediction. 
We adopt BCE and IoU losses for supervision.

The rest of this section will introduce PAG with full-pixel hyperassociation and the condensation of hyperassociations by plugging the Correspondence-induced Association Condensation (CAC) module into PAG.

% \vspace{-2mm}
\subsection{Progressive Association Generation}
% \vspace{-1mm}
Our deep association learning involves two steps: 1) acquiring the raw hyperassociations $\{\boldsymbol{A}_s\}_{s=3}^{5}$ from three stages; 2) employing aggregation networks on $\{\boldsymbol{A}_s\}_{s=3}^{5}$ to obtain association features $\{\boldsymbol{F}_s^{A}\}_{s=3}^{5}$.

Earlier methods \cite{cho2021cats, hong2022cost, shi2023videoflow} directly utilize the original backbone features, \ie $\mathbb{F}^{P}$, to calculate inter-image interactions, such as the so-called cost volume. We argue that hyperassociations derived straight from backbone features might obstruct further improvement of deep association learning, given that the current backbone is pre-trained without any consideration for inter-image associations. 

% Upon this consideration, 
To alleviate this problem, we propose the PAG module to progressively generate pyramid association features so that we can utilize the high-level association feature, \eg $\boldsymbol{F}_{s+1}^{A}$ from the ($s+1$)-th stage, to enhance the neighbouring low-level VGG features in $\mathbb{F}^{P}$, \eg $\boldsymbol{F}_{s,l}$, for attaining association-enhanced features $\hat{\boldsymbol{F}}_{s,l}$, based on which we can calculate high-quality hyperassociations $\boldsymbol{A}_{s}$ in the $s$-th stage. After that, we execute the subsequent aggregation networks on $\boldsymbol{A}_{s}$ to achieve association features $\boldsymbol{F}_{s}^{A}$, which will continue to enhance the VGG features of the next stage and carry out progressive association generation. The whole process of our PAG can be formulated as follows:
\begin{equation}
\begin{gathered}
\boldsymbol{A}_{s} = \operatorname{HAC}\big(\{\hat{\boldsymbol{F}}_{s,l}\}_{l=1}^{L_s}\big), \\
\boldsymbol{F}_{s}^{A} = \operatorname{Agg}\big(\boldsymbol{A}_s\big), \\
\{\hat{\boldsymbol{F}}_{s-1,l}\}_{l=1}^{L_{s-1}} = \operatorname{Enh}\big(\{\boldsymbol{F}_{s-1,l}\}_{l=1}^{L_{s-1}}; \boldsymbol{F}_{s}^{A}\big),
\end{gathered}
\end{equation}
where $s$ ranges from 5 to 3 and $\{\hat{\boldsymbol{F}}_{5,l}\}_{l=1}^{L_5}=\{\boldsymbol{F}_{5,l}\}_{l=1}^{L_5}$. The $\operatorname{HAC}$, $\operatorname{Agg}$, and $\operatorname{Enh}$ represent the hyperassociation calculation, aggregation network, and association-induced feature enhancement, respectively. Next, we explain them in detail.

\Paragraph{Hyperassociation Calculation.} 
For each stage, we first compute the raw associations at each layer using the inner product between $l$-2 normalized association-enhanced features of $N$ related images. After that, we stack the raw associations of all layers to form the final hyperassociation of this stage. The hyperassociation for the $s$-th stage, \ie $\boldsymbol{A}_s \in \mathbb{R}^{N \times H_s \times W_s \times N \times H_s \times W_s \times L_s}$, can be calculated via:  
\begin{equation}
\label{HAC}
\begin{aligned}
\boldsymbol{A}_{s} &= \operatorname{HAC}\big(\{\hat{\boldsymbol{F}}_{s,l}\}_{l=1}^{L_s}\big), \\
&= \operatorname{Stack}\Big(\big\{\operatorname{ReLU}\big(\frac{\hat{\boldsymbol{F}}_{s,l} \cdot \hat{\boldsymbol{F}}_{s,l}^{\top}}{\|\hat{\boldsymbol{F}}_{s,l}\|\|\hat{\boldsymbol{F}}_{s,l}^{\top}\|}\big)\big\}_{l=1}^{L_s}\Big),
\end{aligned}
\end{equation}
where  $\hat{\boldsymbol{F}}_{s,l} \cdot \hat{\boldsymbol{F}}_{s,l}^{\top} \in \mathbb{R}^{N \times H_s \times W_s \times N \times H_s \times W_s}$.  The $\top$ indicates transposing the last dimension and the first three dimensions. $\| \cdot \|$ represents the $l$-2 norm. We employ $\operatorname{ReLU}$ to suppress noisy association values.

\label{aggregation_section}
\Paragraph{Aggregation Network.} The raw hyperassociation $\boldsymbol{A}_s \in \mathbb{R}^{N\times H_s\times W_s\times\textcolor{gray}{N\times H_s\times W_s}}$ $\mathcal{}^{\textcolor{gray}{\times L_s}}$ is a hypercube with a nested structure, where each pixel position is characterized by a 4D tensor ($\textcolor{gray}{N\times H_s\times W_s\times L_s}$). Each 4D tensor documents the associations of the respective pixel with all other pixels in $N$ related images. For clarity, we designate the first and second $N\times H_s\times W_s$ dimensions in $\boldsymbol{A}_s$ as the source and target dimensions, respectively. Although these 4D tensors are crucial for exploring the consensus information for co-saliency detection, they essentially comprise pixel-to-pixel similarity values, seen in (\ref{HAC}), which may be suboptimal and unreliable in complex scenarios. Therefore, we propose \textit{using deep networks} to transform these pixel-wise similarities into deep association features with contextual and high-order association knowledge. This has never been explored in previous CoSOD methods. This is implemented via context aggregations on $\boldsymbol{A}_s$ to squeeze these 4D tensors as $C_s$-dimensional vectors, formulated as:
\begin{equation}
N \! \times \! H_s \! \times \! W_s \! \times \! \textcolor{gray}{N \! \times \! H_s \! \times \! W_s \! \times \! L_s}
\rightarrow N \! \times \! H_s \! \times \! W_s \! \times \! \textcolor{gray}{C_s}.
\end{equation}

In detail, we first deploy several association aggregation layers on $\boldsymbol{A}_s$ to progressively aggregate context information, enlarging $\textcolor{gray}{L_s}$ as $\textcolor{gray}{C_s}$, and eliminate the target $\textcolor{gray}{H_s \times W_s}$ dimension in 4D tensors. Each aggregation layer consists of 2D convolution layers and a downsampling operation. Specifically, focusing on the first aggregation layer for technical explanation, we first aggregate context information in the target $\textcolor{gray}{H_s \times W_s}$ dimension by applying a 2D convolution layer on all 4D tensors. The operations on the 4D tensor at pixel position $(h_i, w_i)$ in image $I_i$ can be formulated as:
% \vspace{-1mm}
\begin{equation}
\begin{gathered}
\label{conv1}
\boldsymbol{A}_s^{1}(i, h_i, w_i, j, :, :, :) = \mathcal{C}_{1}(\boldsymbol{A}_s(i, h_i, w_i, j, :, :, :)), \quad j=1, \cdots, \textcolor{gray}{N},
\end{gathered}
% \vspace{-1mm}
\end{equation}
where $\mathcal{C}_1$ is a $3\times 3$ 2D convolution layer. Here, $j$ is the index of other related images in the 4D tensor, and $\boldsymbol{A}_s(i,h_i,w_i,j,:,:,:) \in \mathbb{R}^{\textcolor{gray}{H_s \times W_s \times L_s}}$ illustrates the associations between the pixel $(h_i,w_i)$ in $I_i$ and all pixels in image $I_j$. This interpretation applies to other similar symbols. 

Then, a downsampling operation $\mathcal{D}$, \ie bilinear interpolation, is applied on $\boldsymbol{A}_s^1$ to reduce the spatial dimension of the 4D tensor by a scaling factor:
\begin{equation}
\begin{gathered}
\label{downsampling}
\boldsymbol{A}_s^{2}(i, h_i, w_i, j, :, :, :) = \mathcal{D}(\boldsymbol{A}_s^{1}(i, h_i, w_i, j, :, :, :)),
\end{gathered}
\end{equation}
where $\boldsymbol{A}_s^{2}(i, h_i, w_i, j, :, :, :) \in \mathbb{R}^{\textcolor{gray}{H_s^{\prime}\times W_s^{\prime}\times C_s^{\prime}}}$, 
$\textcolor{gray}{H_s^{\prime}}$ and  $\textcolor{gray}{W_s^{\prime}}$ are downsampled height and width. $\textcolor{gray}{C_s^{\prime}}$ is the channel number after convolution $\mathcal{C}_{1}$.

Finally, we also aggregate context information in the source $H_s\times W_s$ dimension. Specifically, we extract the values along the source dimension and channel dimension in $\boldsymbol{A}_s^2$ to form 4D tensors and apply a 2D convolution layer on them. For instance, $\boldsymbol{A}_s^2(:,:,:,j,h_j,w_j,:)\in \mathbb{R}^{N\times H_s\times W_s\times \textcolor{gray}{C_s^{\prime}}}$ is such a 4D tensor, where $(h_j,w_j)$ is a pixel position in the target dimension. This can be formulated as:
\begin{equation}
\begin{gathered}
\label{conv2}
\boldsymbol{A}_s^{3}(i, :, :, j, h_j, w_j, :) = \mathcal{C}_{2}\big(\boldsymbol{A}_s^{2}(i, :, :, j, h_j, w_j, :)\big), \quad i=1, \cdots, N,
\end{gathered}
\end{equation}
where $\mathcal{C}_2$ is a $3\times 3$ 2D convolution layer. $i$ is the related image index.

After several association aggregation layers, as shown in (\ref{conv1})-(\ref{conv2}), we can achieve the aggregated association features with the target $\textcolor{gray}{H_s\times W_s}$ dimension eliminated, denoted as $\boldsymbol{F}_{s}^{A^{\prime}} \in \mathbb{R}^{N\times H_s\times W_s\times \textcolor{gray}{N}\times C}$. Subsequently, we average $\boldsymbol{F}_{s}^{A^{\prime}}$ on its second $\textcolor{gray}{N}$ dimension and obtain the final association features $\boldsymbol{F}_s^{A}\in \mathbb{R}^{N\times H_s\times W_s\times C}$, formulated as:
\begin{equation}
\begin{gathered}
\boldsymbol{F}_s^{A} = \frac{1}{N}\mathop{\small\sum}\nolimits_{j=1}^{N}\boldsymbol{F}_{s}^{A^{\prime}}(:, :, :, j, :).
\end{gathered}
\end{equation}

\Paragraph{Association-induced Feature Enhancement.} Once we have obtained association feature $\boldsymbol{F}_s^A$ of the $s$-th stage, we will use it to enhance the VGG feature of the $(s\!-\!1)$-th stage, \ie $\{\boldsymbol{F}_{s-1,l}\}_{l=1}^{L_{s-1}}$. Specifically, we upsample $\boldsymbol{F}_s^A$ to align the spatial size of features in the $(s\!-\!1)$-th stage, and then add it to $\{\boldsymbol{F}_{s-1,l}\}_{l=1}^{L_{s-1}}$ followed by a convolution layer, formulated as:
\begin{equation}
\hat{\boldsymbol{F}}_{s-1,l} = \mathcal{C}_3\big(\boldsymbol{F}_{s-1,l} + \mathcal{U}(\boldsymbol{F}_s^A)\big),
\end{equation}
where $\mathcal{C}_3$ and $\mathcal{U}$ represent a 2D convolution layer and the bilinear upsampling operation, respectively.

% \vspace{-2mm}
\subsection{Correspondence-induced Association Condensation}
% \vspace{-1mm}
Although PAG based on full-pixel hyperassociations can deliver satisfactory performance in CoSOD, it also introduces substantial computational overhead. Additionally, we argue that for each pixel in an image, it is unnecessary to gather its associations with all pixels of other related images to form hyperassociations.
Some pixel associations may even impair the final performance, such as those between ambiguous regions. To this end, this subsection try to condense the original full-pixel hyperassociations to retain only informative pixel associations.

\begin{figure}[t]
\centering
\includegraphics[width=0.9\linewidth]{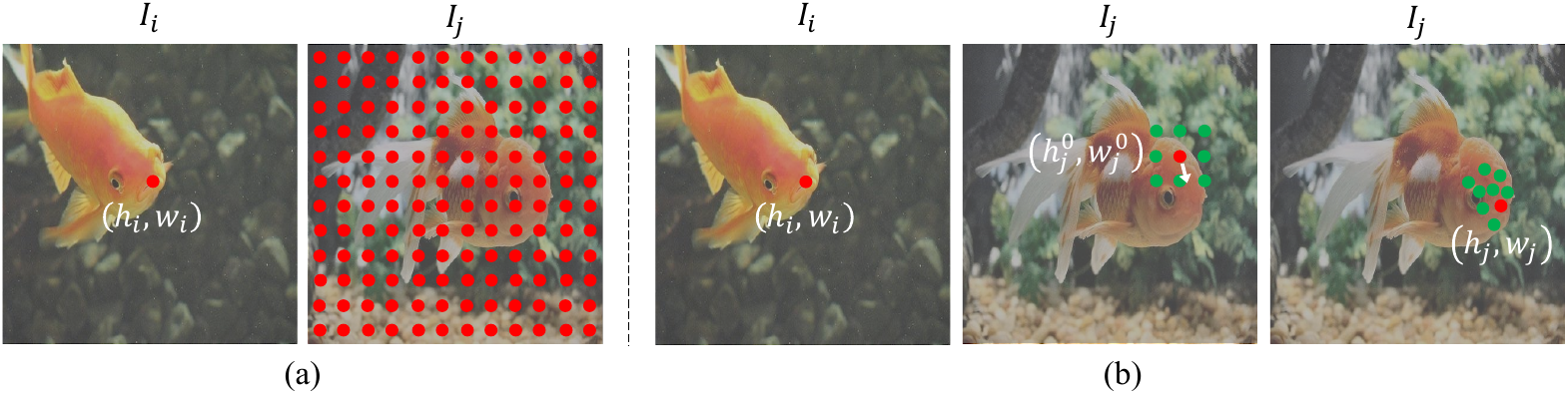}
\vspace{-3mm}
\caption{\textbf{Difference between full-pixel hyperassociation (a) and condensed hyperassociation (b).} 
We provide an example of collecting the pixel associations from image $I_j$ for a pixel $(h_i, w_i)$ in image $I_i$.
Full-pixel hyperassociation collects all pixel associations in $I_j$, while our condensed hyperassociation only collects the associations of its correspondence pixel $(h_j, w_j)$ (red dot) and surrounding pixels (green dots). We first heuristically find an initial pixel $(h_j^0, w_j^0)$ with a fixed surrounding window and then learn coordinate offsets to locate the optimized correspondence and surrounding pixels.}
\label{correspondence_explanation}
\vspace{-6mm}
\end{figure}

This subsection will focus on explaining the condensation of pixel associations of a pixel (\eg $(h_i, w_i)$ in $I_i$) to ones of the other images (\eg image $I_j$), \ie $\boldsymbol{A}_s(i,h_i,w_i,j,:,:,:) \in \mathbb{R}^{H_s\times W_s\times L_s}$, as shown in Figure~\ref{correspondence_explanation}. We will simplify the symbol $\boldsymbol{A}_s(i,h_i,w_i,j,:,:,:)$ as $\boldsymbol{a}_s^j$ in the subsequent text for convenience.

Specifically, CAC opts to select $K \!\times\! K$ ($K\!\!<\!\!H_s,K\!\!<\!\!W_s$) informative pixel associations from $\boldsymbol{a}_s^j$ to form its condensed representation, \ie $\tilde{\boldsymbol{a}}_s^j \in \mathbb{R}^{K\times K\times L_s}$. Thus, the entire condensed hyperassocaiton can be symbolized as $\tilde{\boldsymbol{A}}_s \in \mathbb{R}^{N\times H_s \times W_s \times N \times}$ $\mathbb{}^{K \times K \times L_s}$. To ensure the proper selection of $K \!\times\! K$ pixels, we introduce a pretext task, \ie semantic correspondence estimation \cite{liu2020semantic, kim2022transformatcher}. This allows us to first locate the corresponding pixel of $(h_i, w_i)$ in image $I_j$, \ie $(h_j, w_j)$, and then combine $(h_j, w_j)$ with its surrounding pixels as the $K \!\times\! K$ pixel set.
We design this approach based on the observation that the co-salient objects across $N$ related images belong to the same semantic category, and the pixels within them should have semantic correspondences to each other, as shown in Figure~\ref{Correspondence_Visual}. Therefore, the introduction of semantic correspondence in CAC not only improves the CoSOD performance but also delves deeper into the core nature of the CoSOD task. As far as we know, this is the first work to use semantic correspondence in the CoSOD task.

\Paragraph{Correspondence Estimation.} 
To estimate the correspondence pixel $(h_j, w_j)$ in $I_j$ for $(h_i, w_i)$, we first identify an initial pixel $(h_j^0, w_j^0)$ via a heuristic approach. Subsequently, we produce a spatial offset to refine $(h_j^0, w_j^0)$ into $(h_j, w_j)$. To achieve this purpose, all initial correspondence pixels should be utilized to form the initial condensed hyperassociations, with which the initial deep association features can be generated for spatial offset prediction.

Specifically, we pick out $(h_j^0, w_j^0)$ that has the largest feature similarity value with $(h_i, w_i)$. Since we have calculated the feature similarities in $\boldsymbol{a}_s^j$, the $(h_j^0, w_j^0)$ can be obtained as follows:
\begin{equation}
\begin{aligned}
\boldsymbol{\bar{a}}_s^j &= \mathop{\sum}\nolimits_{l=1}^{L_s} \boldsymbol{a}_s^j(:,:,l), \\
(h_j^0, w_j^0) &= \underset{x,y}{\operatorname{argmax}}(\boldsymbol{\bar{a}}_s^j(x,y)),
\end{aligned}
\end{equation}
where $\boldsymbol{\bar{a}}_s^j \in \mathbb{R}^{H_s \times W_s}$ is a similarity matrix obtained by eliminating the last dimension of $\boldsymbol{a}_s^j \in \mathbb{R}^{H_s \times W_s \times L_s}$ through summation. $\operatorname{argmax}$ returns the coordinate of the maximum value.

Next, we select $K\times K$ pixels within the square region centered around the initial pixels, \eg $(h_j^0, w_j^0)$, to construct the initial condensed hyperassociation $\tilde{\boldsymbol{A}}_s^{\prime}$. Then, we can feed it into the aggregation network, described in Sec~\ref{aggregation_section}, and achieve the initial aggregated association features $\boldsymbol{F}_s^{A^{\prime}} \in \mathbb{R}^{N\times H_s\times W_s\times N\times C_s}$. It can be regarded as the association features of each pixel to $N$ other related images. We utilize the feature of $(h_i,w_i)$ to $I_j$, 
\ie $\boldsymbol{F}_s^{A^{\prime}}(i,h_i,w_i,j,:)\in \mathbb{R}^{C_s}$, to predict the offsets for $(h_j, w_j)$ via a linear layer, formulated as:
\begin{equation}
\begin{gathered}
\boldsymbol{\Delta}_s^j = \mathcal{O}(\boldsymbol{F}_s^{A^{\prime}}(i,h_i,w_i,j,:)),
\end{gathered}
\end{equation}
where $\mathcal{O}$ is a linear layer for offset generation. $\boldsymbol{\Delta}_s^j \in \mathbb{R}^{K\times K\times 2}$ consists of $K\!\times\!K$ offsets, besides the center offset $\boldsymbol{\Delta}_s^j(k_c,k_c,:)$ for correspondence estimation, \ie refining $(h_j^0, w_j^0)$ as $(h_j, w_j)$, we also generate other offsets for surrounding pixel selection. Thus, the corresponding pixel $(h_j, w_j)$ can be obtained by adding the offset to the initial pixel $(h_j^0, w_j^0)$, formulated as:
\begin{equation}
(h_j, w_j) = (h_j^0, w_j^0) +  \boldsymbol{\Delta}_s^j(k_c, k_c, :),
\end{equation}
where $(k_c, k_c)$ is the center position of $K\!\times\!K$ square.

\Paragraph{Condensation Operation.} Given the estimated correspondence pixel $(h_j, w_j)$ and other offsets in $\boldsymbol{\Delta}_s^j$, we can obtain the surrounding pixels and combine them with $(h_j, w_j)$ to form $K\!\times\!K$ pixel set. Their coordinates are stored in $\boldsymbol{n}_s^j \in \mathbb{R}^{K\times K\times 2}$. This process can be formulated as:
\begin{equation}
\begin{aligned}
&\boldsymbol{n}_s^j(x, y, :) = (h_j, w_j) + \boldsymbol{\Delta}_s^j(x, y, :), \quad x,y \in \{1, \cdots, K\};\,x,y \neq k_c, \\
&\boldsymbol{n}_s^j(k_c, k_c, :) = (h_j, w_j),
\end{aligned}
\end{equation}
where $(k_c, k_c)$ is the center position of $K\!\times\!K$ square.

Finally, we can perform the condensation operation via the index selection on $\boldsymbol{a}_s^j$, formulated as:
\begin{equation} 
\begin{gathered} 
\tilde{\boldsymbol{a}}_s^j = \boldsymbol{a}_s^j( \boldsymbol{n}_s^j),
\end{gathered}
\end{equation}
where $\tilde{\boldsymbol{a}}_s^j$ is the condensed representation of $\boldsymbol{a}_s^j$, \ie pixel associations of $(h_i, w_i)$ to image $I_j$. Furthermore, we also illustrate the condensation for pixel associations of $(h_i, w_i)$ to all images, \ie $\boldsymbol{a}_s \in \mathbb{R}^{N\times H_s\times W_s\times L_s}$, in Figure~\ref{framework}. By applying such a condensation process to all pixel associations, we can obtain the final condensed hyperassociation $\tilde{\boldsymbol{A}}_s$.

% \vspace{-3mm}
\subsection{Object-aware Cycle Consistency Loss} 
% \vspace{-1mm}
To achieve accurate correspondence estimations, applying effective supervisions on them is necessary. As explicit semantic correspondence annotations are not available, we can only rely on unsupervised losses by imposing correspondence-related constraints on estimated correspondences. Previous unsupervised approaches apply constraints to all pixels \cite{shen2020ransac, truong2021warp, zhao2021multi}, including those on the background and extraneous objects that don't have mutual correspondences, hence harming the model effectiveness. To avoid this problem, we propose an object-aware constraint to only access losses on the co-salient pixels.

We propose an Object-aware Cycle Consistency (OCC) loss for the supervision of correspondence estimations in CoSOD. The cycle consistency can be explained as: if a co-salient pixel $(h_i, w_i)$ of image $I_i$ corresponds to the pixel $(h_j, w_j)$ in image $I_j$, 
then the pixel $(h_j, w_j)$ should also semantically correspond to pixel $(h_i, w_i)$. 

Based on this cycle consistency constraint, we adopt image warping operations to conduct the OCC loss. Specifically, we first warp the image $I_i^s$ 
(a resized $I_i$ to align the scales in the $s$-th stage) 
as $I_{i\rightarrow j}^s$ using the $I_i^s\rightarrow I_j^s$ correspondence estimations.
Next, we warp $I_{i\rightarrow j}^s$ backward to $I_{i\rightarrow j \rightarrow i}^s$
using $I_j^s\rightarrow I_i^s$ correspondence estimations.
Finally, we can utilize the SSIM loss between $I_i^s$ and $I_{i\rightarrow j \rightarrow i}^s$ to measure the cycle consistency for mutual correspondence pixels in $I_i^s$ and $I_j^s$. 
Moreover, to ensure the constraints are only conducted on co-salient objects, we mask the images with their ground truth masks, formulated as:
\vspace{-2mm}
\begin{equation}
\begin{gathered}
\mathcal{L}_s^{C} = \frac{1}{N^2}\sum_{i=1}^{N}\sum_{j=1}^{N}\mathcal{L}_{SSIM}(I_i^s\cdot G_i^s, I_{i\rightarrow j\rightarrow i}^s\cdot G_i^s),
\end{gathered}
\vspace{-2mm}
\end{equation}
where $G_i^s$ is the resized ground truth of image $I_i$. The total OCC loss $\mathcal{L}^{C}$ is the sum of three stages, formulated as: $\mathcal{L}^{C}=\mathcal{L}_3^C + \mathcal{L}_4^C + \mathcal{L}_5^C$. More details can be found in the supplementary materials.

% \vspace{-1mm}
\section{Experiments}
% \vspace{-2mm}
\subsection{Evaluation Datasets and Metrics}
% \vspace{-1mm}
We follow \cite{fan2021GCoNet} to evaluate our model on three benchmark datasets, \ie CoCA \cite{zhang2020gicd} (1295 images of 80 groups), CoSal2015 \cite{zhang2016detection} (2015 images of 50 groups), and CoSOD3k \cite{fan2021re} (3316 images of 160 groups). 
% CoCA \cite{zhang2020gicd} is the most challenging dataset and contains 1295 images organized into 80 groups. CoSal2015 \cite{zhang2015co} consists of 50 groups with a total of 2015 images, while CoSOD3k \cite{fan2020taking} is made up of 160 groups with 3316 images. 
We adopt four widely-used metrics for the quantitative evaluation, \ie Structure-measure ($S_m$) \cite{fan2017structure}, Maximum enhanced-alignment measure ($E_\xi$) \cite{Fan2018Enhanced}, Maximum F-measure ($\text{F}_\beta$) \cite{5206596}, and Mean Absolute Error ($\mathcal{M}$) \cite{6751300}.

% \vspace{-3mm}
\subsection{Implementation Details}
% \vspace{-1mm}
To construct the training data, we follow \cite{zheng2023gconet+} to use different combinations of three commonly used training datasets, \ie DUTS class \cite{zhang2020gicd} (8250 images of 291 groups), COCO-9k \cite{lin2014microsoft} (9213 images of 65 groups), and COCO-SEG \cite{wang2019robust} (200,000 images of 78 groups), for a fair comparison with other state-of-the-art (SOTA) works. We also implement the synthesis strategy for the DUTS class dataset following \cite{zhang2021summarize}. 

For training specifics, we employ the data augmentation strategy in \cite{liu2018picanet} and use $256\times 256$ as the input size for the network. We employ the Adam optimizer \cite{kingma2014adam} with $\beta_1=0.9$ and $\beta_2=0.99$ to optimize the network. We train our CONDA model for 300 epochs, starting with an initial learning rate of $1e-4$, which is divided by 10 at the $60,000^{th}$ iteration. Our experiments are implemented
based on PyTorch \cite{paszke2019pytorch} on a single Tesla A40 GPU, with the batchsize set to $N=6$. The hyperparameter $K$ in CAC is set to 9.

\begin{table}[t]
\centering
\scriptsize
% \footnotesize
\renewcommand{\arraystretch}{0.9}
\renewcommand{\tabcolsep}{1.0mm}
\caption{\textbf{Ablation Study of our proposed modules.} 
% Seperate Association Generation (SAG), Similarity-induced Association Condensation (SAC), and Full-pixel Cycle Consistency (FCC) are ablation modules for PAG, CAC, and OCC, respectively.
SAG, SAC, and FCC are ablation modules for PAG, CAC, and OCC, respectively.
}
\vspace{-3mm}
\begin{tabular}{c|cccccc|cccc}
\toprule

\multicolumn{1}{c|}{\multirowcell{2}{ID}} &
\multicolumn{6}{c|}{Modules} & 
\multicolumn{4}{c}{CoCA \cite{zhang2020gicd}} 
\\

\cmidrule{2-11}

&
\multicolumn{1}{c}{SAG} &
\multicolumn{1}{c|}{PAG} &
\multicolumn{1}{c}{SAC} &
\multicolumn{1}{c|}{CAC} &
\multicolumn{1}{c}{FCC} &
\multicolumn{1}{c|}{OCC} &
\multicolumn{1}{c}{$S_m \uparrow$} &  
\multicolumn{1}{c}{$E_\xi \uparrow$} &
\multicolumn{1}{c}{$\text{F}_\beta \uparrow$} & 
\multicolumn{1}{c}{$\mathcal{M} \downarrow$}\\ \midrule

1 &
\multicolumn{1}{c}{} &
\multicolumn{1}{c|}{} &
\multicolumn{1}{c}{} &
\multicolumn{1}{c|}{} &
\multicolumn{1}{c}{} &
\multicolumn{1}{c|}{} &
0.6936 & 0.7642 & 0.5729 & 0.1206 
\\ \midrule

2 &
\multicolumn{1}{c}{\checkmark} &
\multicolumn{1}{c|}{} &
\multicolumn{1}{c}{} &
\multicolumn{1}{c|}{} &
\multicolumn{1}{c}{} &
\multicolumn{1}{c|}{} &
0.7236 & 0.8029 & 0.6357 & 0.1106 
\\ 

\rowcolor[RGB]{214,220,229}
3 &
\multicolumn{1}{c}{} &
\multicolumn{1}{c|}{\checkmark} &
\multicolumn{1}{c}{} &
\multicolumn{1}{c|}{} &
\multicolumn{1}{c}{} &
\multicolumn{1}{c|}{} &
0.7308 & 0.8122 & 0.6459 & 0.1075 
\\ \midrule

4 &
\multicolumn{1}{c}{} &
\multicolumn{1}{c|}{\checkmark} &
\multicolumn{1}{c}{\hspace{2mm}\checkmark} & 
\multicolumn{1}{c|}{} & 
\multicolumn{1}{c}{} &
\multicolumn{1}{c|}{} & 
0.7304 & 0.8123 & 0.6500 & 0.1085
\\ 

\rowcolor[RGB]{214,220,229}
5 &
\multicolumn{1}{c}{} &
\multicolumn{1}{c|}{\checkmark} &
\multicolumn{1}{c}{} &
\multicolumn{1}{c|}{\checkmark} &
\multicolumn{1}{c}{} &  
\multicolumn{1}{c|}{} &  
0.7473 & 0.8155 & 0.6591 & 0.0956
\\ \midrule

6 &
\multicolumn{1}{c}{} &
\multicolumn{1}{c|}{\checkmark} &
\multicolumn{1}{c}{} &
\multicolumn{1}{c|}{\checkmark} &
\multicolumn{1}{c}{\checkmark} &  
\multicolumn{1}{c|}{} &  
0.7398 & 0.8138 & 0.6506 & 0.1013
\\ 

\rowcolor[RGB]{214,220,229}
7 &
\multicolumn{1}{c}{} &
\multicolumn{1}{c|}{\checkmark} &
\multicolumn{1}{c}{} &
\multicolumn{1}{c|}{\checkmark} &
\multicolumn{1}{c}{} &  
\multicolumn{1}{c|}{\checkmark} &   
\textbf{0.7570} & \textbf{0.8248} & \textbf{0.6751} & \textbf{0.0924}
\\ 

\bottomrule

\end{tabular}
\label{ablationTab}
\vspace{-4mm}
\end{table}

% \vspace{-3mm}
\subsection{Ablation Study}
% \vspace{-1mm}
We conduct ablation studies on the most challenging CoCA \cite{zhang2020gicd} dataset. To construct the baseline, we use the FPN \cite{lin2017feature} (with VGG-16 as encoder) as the foundational segmentation network and enhance it with the Region-to-Region correlation module (R2R) \cite{li2023discriminative} to simply capture inter-image connections.
%in a simple way for the CoSOD task. 
Then, as shown in Table~\ref{ablationTab}, we incrementally incorporate PAG, CAC, and OCC into the baseline for effectiveness analysis. We trained all ablation models with the DUTS class and COCO9k datasets.  

\Paragraph{Effectiveness of PAG.} 
% Given intermediate image features from the encoder of the FPN, PAG first utilizes them to calculate full-pixel hyperassociations and then conducts aggregation networks on them to generate deep association features for the subsequent decoder process. 
PAG first uses intermediate image features from the FPN encoder to calculate full-pixel hyperassociations. Then, aggregation networks are applied to generate deep association features for the decoder process.
% Compared to previous image feature optimization strategies, PAG effectively leverages the advantages of deep learning to model pair-wise pixel associations, thereby obtaining higher-level association knowledge. 
PAG effectively utilizes deep learning to model pair-wise pixel associations, achieving higher-level association knowledge compared to previous image feature optimization strategies.
As shown in the 3rd line of Table~\ref{ablationTab}, PAG shows significant performance boosts compared to the baseline model, with respective gains of 3.72\%, 4.80\%, 7.30\%, and 1.31\% in $S_m$, $E_\xi$, $\text{F}_\beta$, and $\mathcal{M}$. 

\begin{figure*}[t]
	\scriptsize
	\renewcommand{\tabcolsep}{0.8pt} % adjust horizontal space
	\renewcommand{\arraystretch}{0.5} % adjust vertical space
	\centering
	\begin{tabular}{cccccc}
	    
		\makecell[c]{\includegraphics[width=0.1051\linewidth,height=0.09\linewidth]{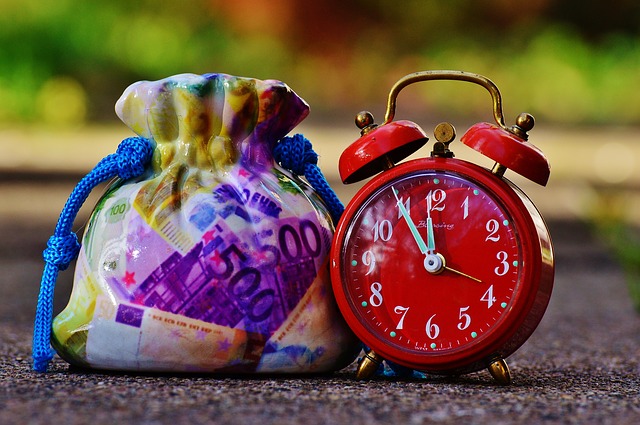}} 
		&
		\makecell[c]{\includegraphics[width=0.1051\linewidth,height=0.09\linewidth]{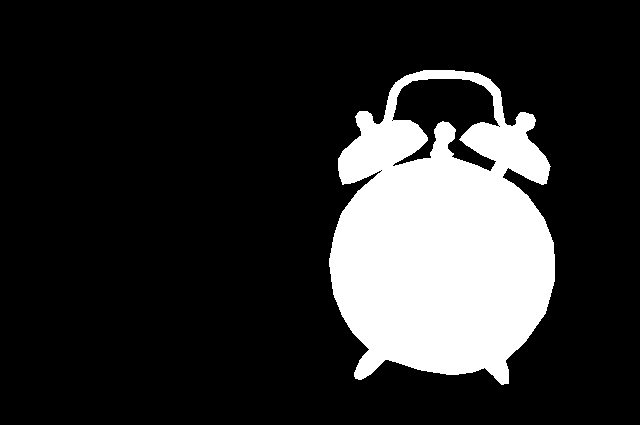}} 
		&
		\makecell[c]{\includegraphics[width=0.1051\linewidth,height=0.09\linewidth]{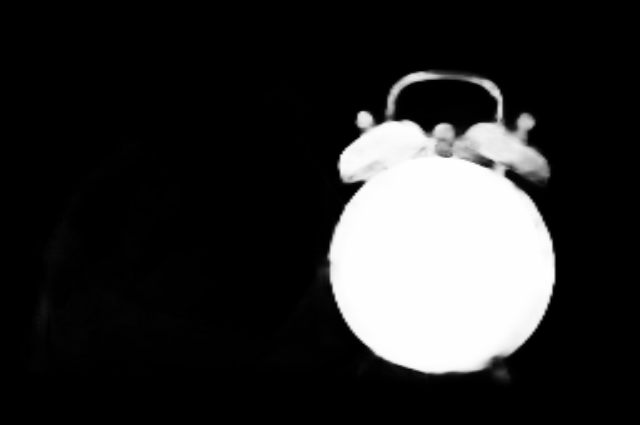}} 
		&
		\makecell[c]{\includegraphics[width=0.22\linewidth,height=0.09\linewidth]{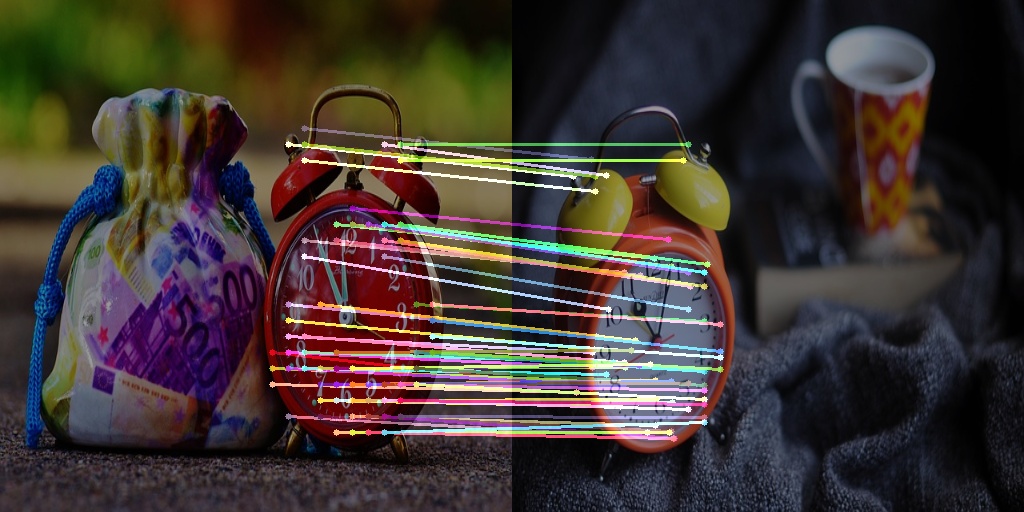}} 
		&
		\makecell[c]{\includegraphics[width=0.22\linewidth,height=0.09\linewidth]{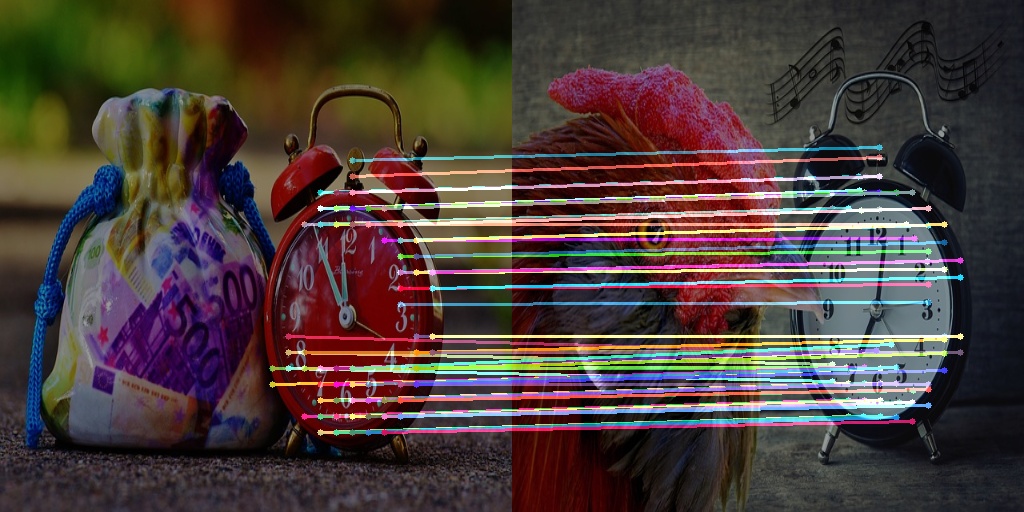}}
            &
		\makecell[c]{\includegraphics[width=0.22\linewidth,height=0.09\linewidth]{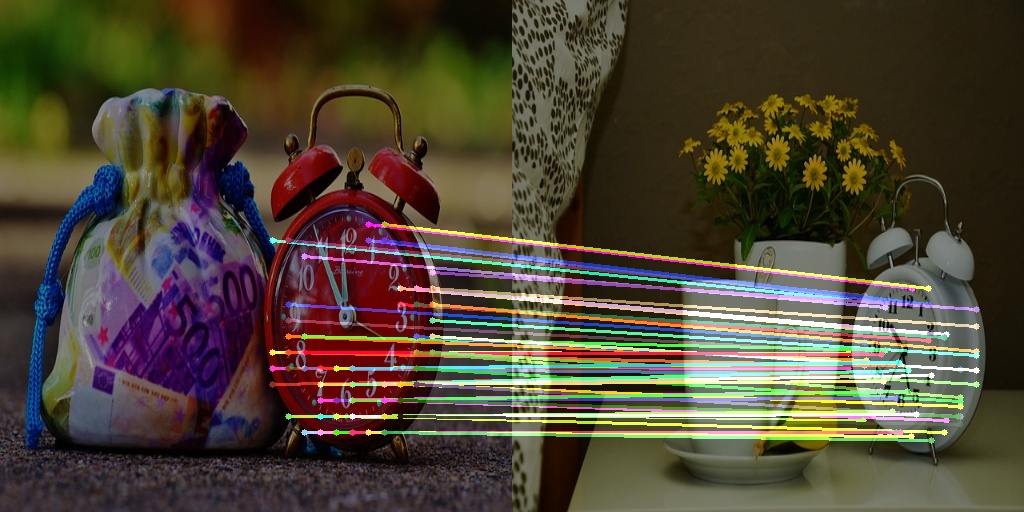}} 
            \vspace{-0.4mm}
            \\
            
		\makecell[c]{\includegraphics[width=0.1051\linewidth,height=0.09\linewidth]{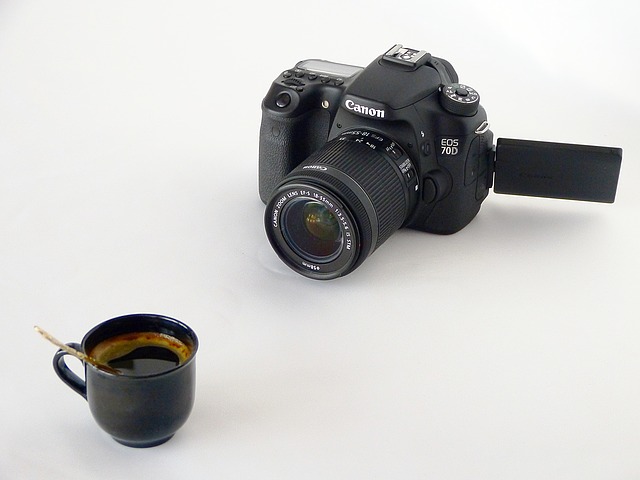}} 
		&
		\makecell[c]{\includegraphics[width=0.1051\linewidth,height=0.09\linewidth]{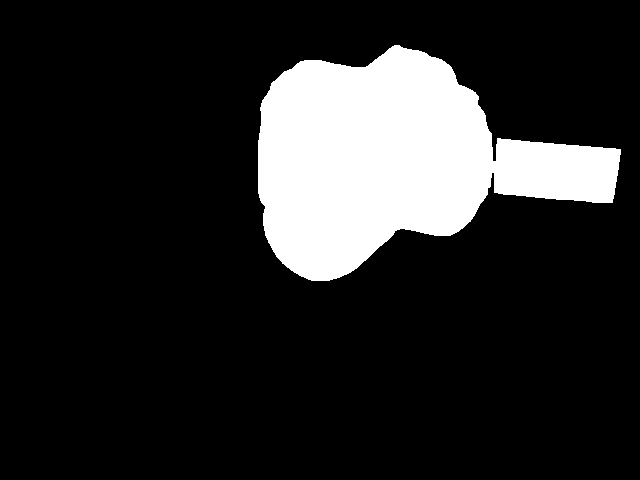}} 
		&
		\makecell[c]{\includegraphics[width=0.1051\linewidth,height=0.09\linewidth]{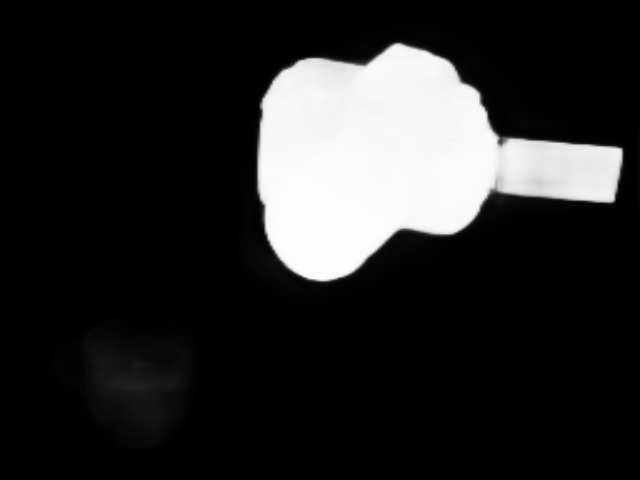}} 
		&
		\makecell[c]{\includegraphics[width=0.22\linewidth,height=0.09\linewidth]{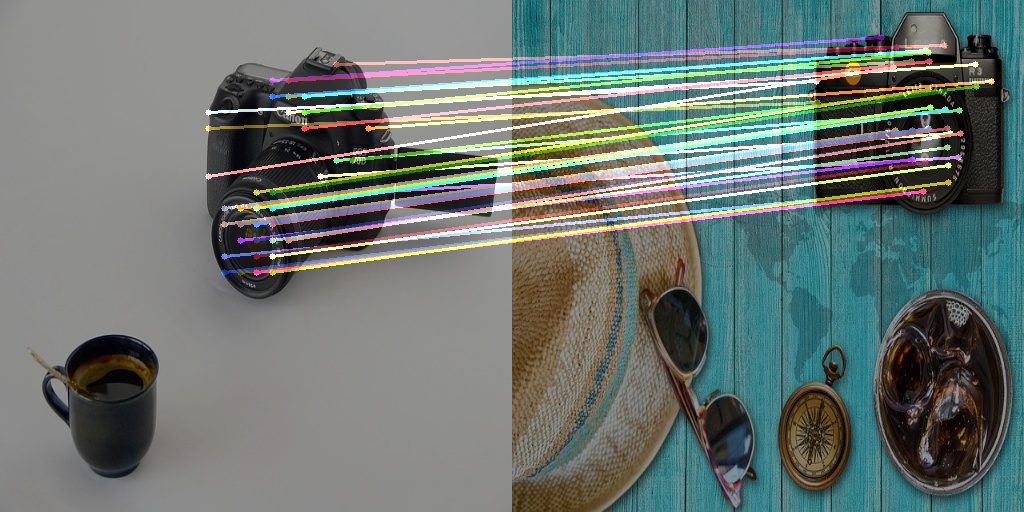}} 
		&
		\makecell[c]{\includegraphics[width=0.22\linewidth,height=0.09\linewidth]{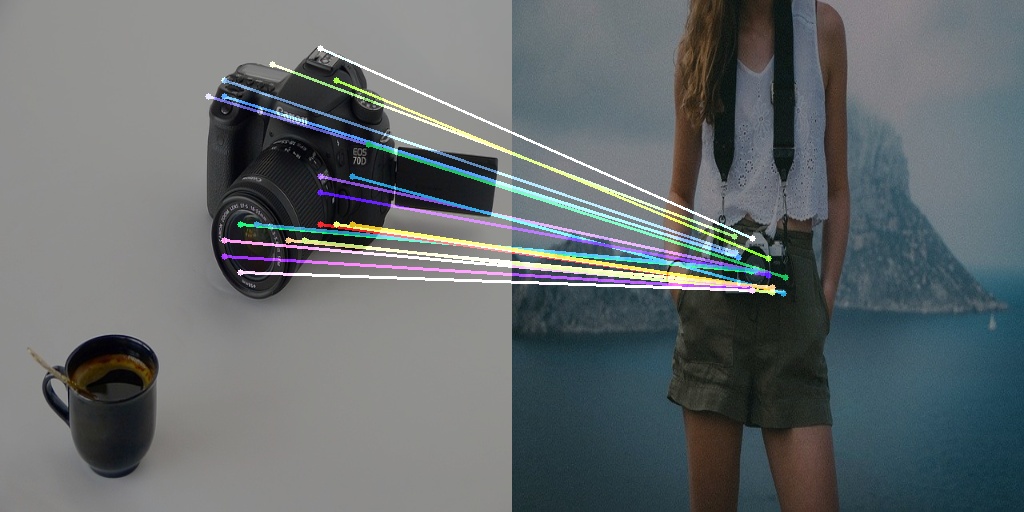}} 
		&
		\makecell[c]{\includegraphics[width=0.22\linewidth,height=0.09\linewidth]{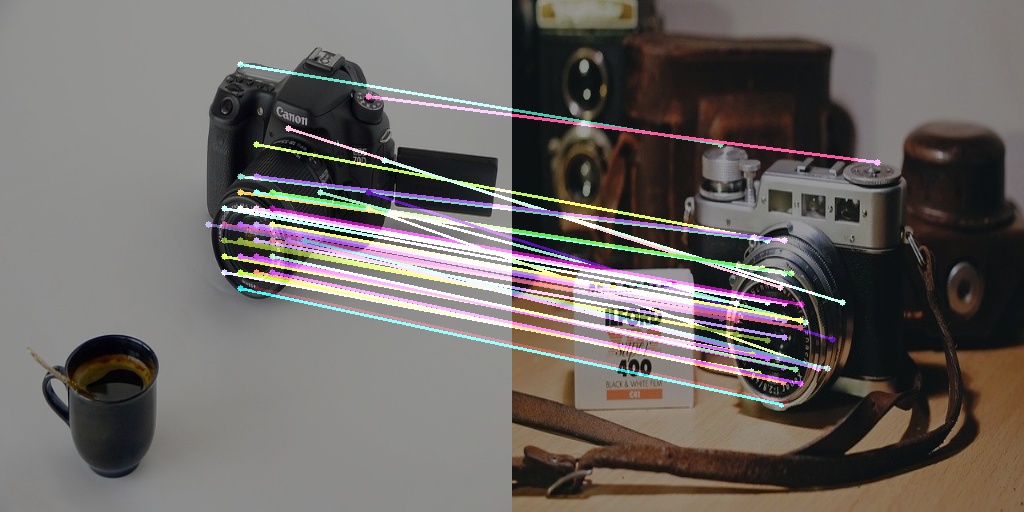}}
            \vspace{-0.4mm}
            \\
            
		\makecell[c]{\includegraphics[width=0.1051\linewidth,height=0.09\linewidth]{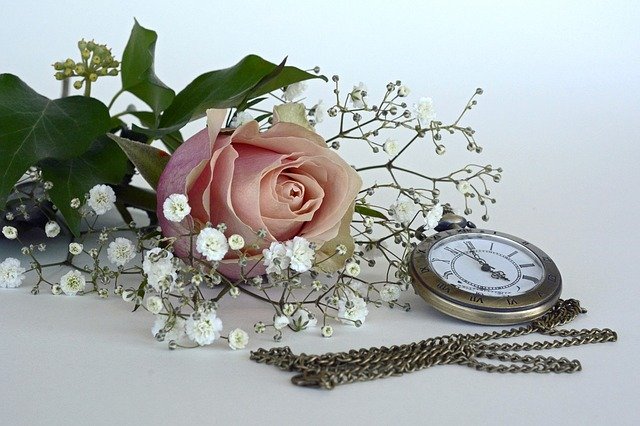}} 
		&
		\makecell[c]{\includegraphics[width=0.1051\linewidth,height=0.09\linewidth]{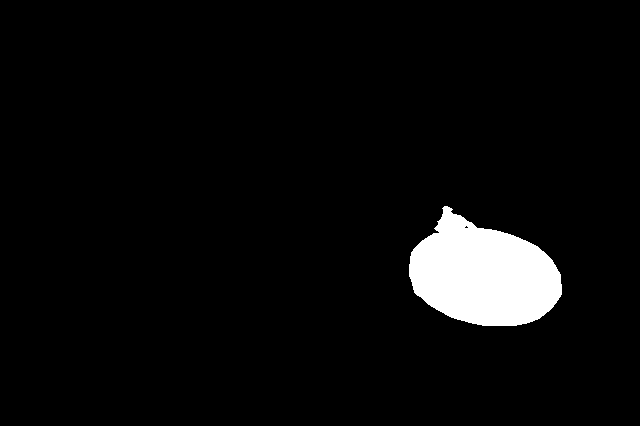}} 
            &
		\makecell[c]{\includegraphics[width=0.1051\linewidth,height=0.09\linewidth]{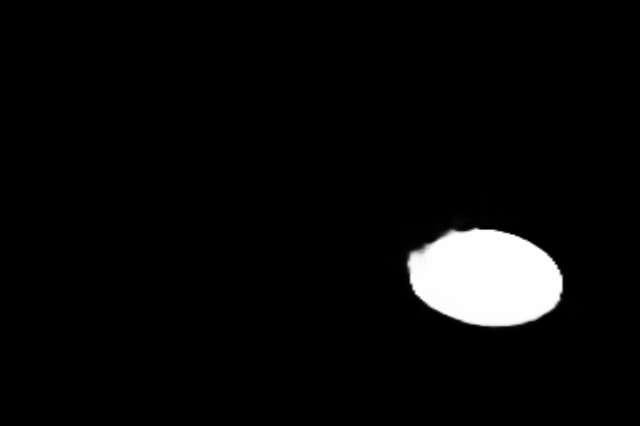}} 
		&
		\makecell[c]{\includegraphics[width=0.22\linewidth,height=0.09\linewidth]{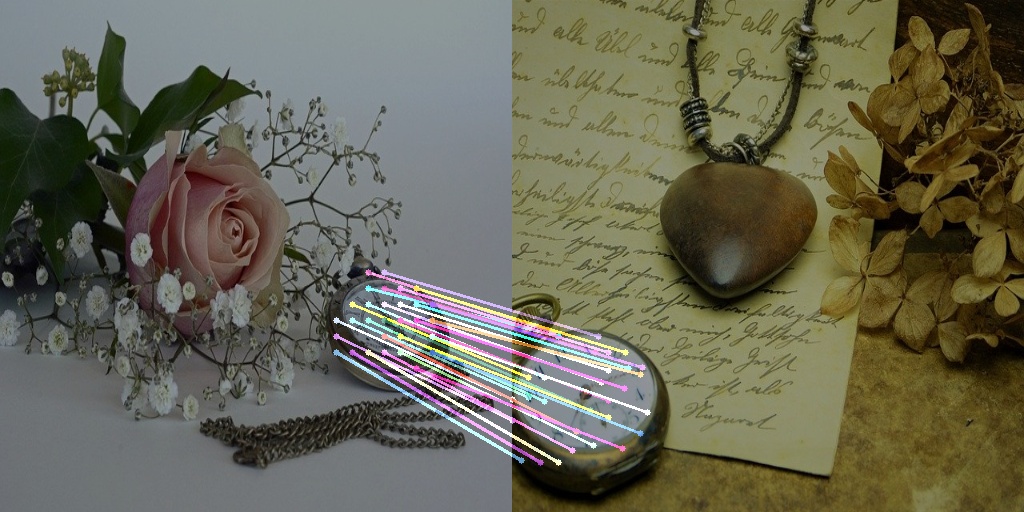}} 
		&
		\makecell[c]{\includegraphics[width=0.22\linewidth,height=0.09\linewidth]{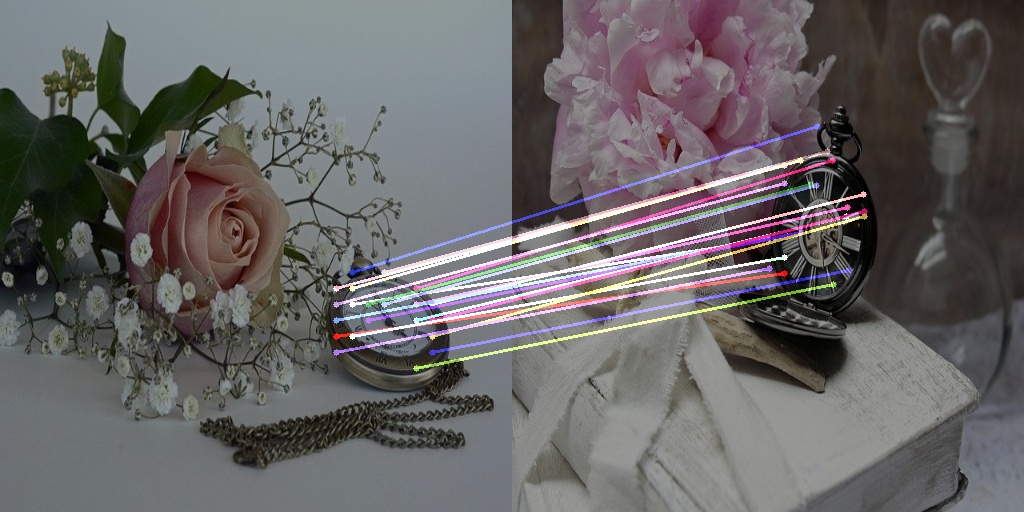}} 
		&
		\makecell[c]{\includegraphics[width=0.22\linewidth,height=0.09\linewidth]{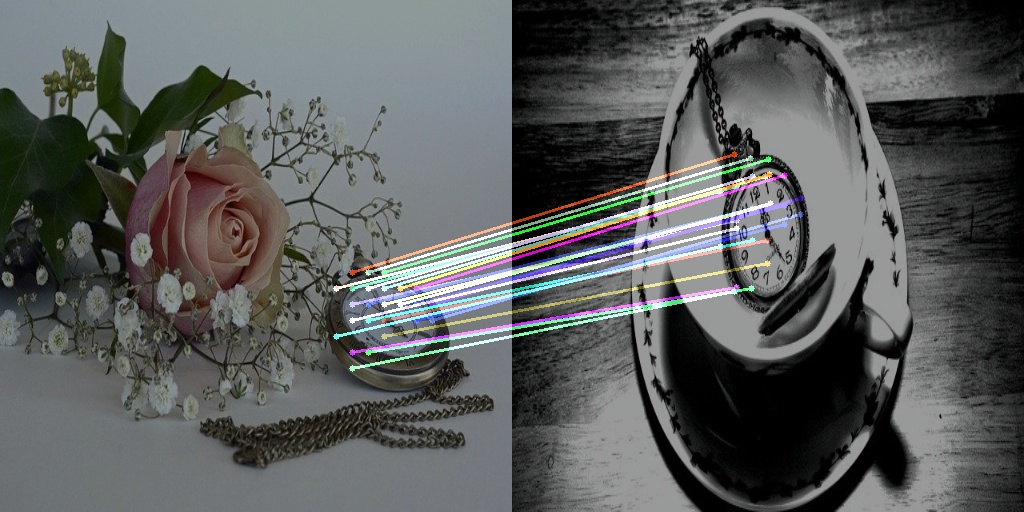}} 
		\\

            Image &
            GT & 
            Prediction &
            Correspondence I &
            Correspondence II &
            Correspondence III
  
		\end{tabular}
            \vspace{-2mm}
		\caption{
		\textbf{Visual samples for the correspondence estimations.} 
  Correspondences I, II, and III visually display estimated correspondences between the main image and three related images. Sparse co-salient pixels were selected and connected to their corresponding pixels using colored lines for clear visualization.
  }
		\label{Correspondence_Visual}
		\vspace{-5mm}
\end{figure*}

Furthermore, to validate our approach of progressively enhancing image features with the previously generated association feature for improving hyperassociation calculation, 
we conduct an ablation experiment called Separate Association Generation (SAG) where association features are generated for three stages without image feature enhancements. 
As shown in the 2nd\&3rd lines in Table~\ref{ablationTab}, PAG outperforms SAG, indicating that our progressive enhancement design can obtain better hyperassociation.

\Paragraph{Effectiveness of CAC.} CAC aims to condense full-pixel hyperassociations in PAG by selecting corresponding pixels and their surrounding contexts. Results in the 3rd\&5th lines of Table~\ref{ablationTab} show that introducing the CAC module improves the performance. Moreover, it reduces the multiply-accumulate operations (MACs) of aggregation networks from 91.38G in the full-pixel PAG to 77.19G\textsuperscript{1}.  This indicates that utilizing correspondence estimation to condense the hyperassociations
not only effectively reduces the computational burden but also helps obtain more accurate pixel associations.

\footnotetext[1]{We input a group of 6 related 256x256 images to measure MACs.}

We also analyze CAC in detail.
As deep association features are necessary for reliable correspondence estimation, CAC first pre-condenses the hyperassociations using a maximum similarity approach to obtain initial deep association features and then performs further condensation based on the correspondences predicted by these initial deep association features.
The hyperassociation condensation process with only the pre-condensation operation is called Similarity-induced Association Condensation (SAC). In Table~\ref{ablationTab}, SAC only brings slight performance improvements due to the heuristic nature of the correspondence estimation. Nevertheless, SAC can provide the initial association feature for CAC to predict reliable correspondence.

\Paragraph{Effectiveness of OCC.}
OCC provides self-supervision for CAC to enable more precise correspondence estimation.
Results in the 5th\&7th lines of Table~\ref{ablationTab} show OCC further improves the performance by leveraging more precise correspondence estimations to condense hyperassociations effectively. 
In addition, we conducted an ablation experiment to validate our object-aware design in CAC by replacing OCC with full-pixel cycle consistency (FCC) loss.
Comparing the 5th\&6th lines of Table~\ref{ablationTab}, FCC leads to a notable performance decrease due to background pixels disrupting the correspondence learning.

\Paragraph{Visualization of Correspondence Estimations.} We present some visual samples of correspondence estimations for some co-salient pixels in Figure~\ref{Correspondence_Visual}. Our semantic correspondence estimations are meaningful and can effectively depict the common attributes of co-salient objects at the pixel level.

\begin{table}[t]
  \centering
  \scriptsize
  \renewcommand{\arraystretch}{1.0}
  \renewcommand{\tabcolsep}{0.4mm}
    \caption{\textbf{Quantitative comparison of our model with other SOTA methods.} $\text{DC}$, $\text{C9}$, and $\text{CS}$ are DUTS class, COCO9k, and COCO-SEG training data, respectively. \textbf{bold} and \underline{underline} mark the best and second-best excellent results, respectively.}
  \vspace{-0.3cm}
  \begin{tabular}{lr|c|cccc|cccc|cccc}
  \toprule
  
   \multicolumn{2}{c|}{\multirow{2}{*}{\raisebox{-0.3em}{Methods}}} &
   \multicolumn{1}{c|}{\multirow{2}{*}{\raisebox{-0.9em}{\makecell{Training \\ Set}}}} &
   \multicolumn{4}{c|}{CoCA\cite{zhang2020gicd}} &
   \multicolumn{4}{c|}{CoSal2015\cite{zhang2016detection}} &
   \multicolumn{4}{c}{CoSOD3k\cite{fan2021re}} 
   \\
   \cmidrule{4-15}

   &
   &
   &
   $S_m \uparrow$ &
   $E_\xi \uparrow$ &
   $\text{F}_\beta \uparrow$ &
   $\mathcal{M} \downarrow$ &
   
   $S_m \uparrow$ &
   $E_\xi \uparrow$ &
   $\text{F}_\beta \uparrow$ &
   $\mathcal{M} \downarrow$ & 
   
   $S_m \uparrow$ &
   $E_\xi \uparrow$ &
   $\text{F}_\beta \uparrow$ &
   $\mathcal{M} \downarrow$ 

   \\
   
   \midrule
   
   \multicolumn{2}{c|}{$\text{GICD}$ 
   } 
   & DC 
   &
   0.658 & 0.718 & 0.513 & 0.126 &
   0.844 & 0.887 & 0.844 & 0.071 &
   0.797 & 0.848 & 0.770 & 0.079 
   \\

   \multicolumn{2}{c|}{$\text{GCoNet}$ 
   } 
   & DC 
   &
   0.673 & 0.760 & 0.544 & \underline{0.110} &
   0.845 & 0.888 & 0.847 & 0.068 &
   0.802 & 0.860 & 0.778 & 0.071 
   \\   

   \multicolumn{2}{c|}{$\text{GCoNet+}$ 
   } 
   & DC 
   &
   \underline{0.691} & \textbf{0.786} & \underline{0.574} & 0.113 &
   \underline{0.875} & \underline{0.918} & \underline{0.876} & \underline{0.054} &
   \underline{0.828} & \textbf{0.881} & \underline{0.807} & \underline{0.068} 
   \\

   \rowcolor[RGB]{214,220,229}
   \multicolumn{2}{c|}{\textbf{$\text{CONDA}$}} 
   & DC 
   &
   \textbf{0.717} & \underline{0.774} & \textbf{0.600} & \textbf{0.102} &
   \textbf{0.890} & \textbf{0.926} & \textbf{0.894} & \textbf{0.049} &
   \textbf{0.832} & \underline{0.873} & \textbf{0.807} & \textbf{0.067}
   \\ 

   \midrule

   \multicolumn{2}{c|}{$\text{ICNet}$ 
   } 
   & C9 
   &
   0.654 & 0.704 & 0.513 & 0.147 &
   \underline{0.857} & 0.901 & \underline{0.858} & \textbf{0.058} &
   0.794 & 0.845 & 0.762 & 0.089 
   \\

   \multicolumn{2}{c|}{$\text{DCFM}$
   } 
   & C9 
   &
   0.710 & 0.783 & 0.598 & \textbf{0.085} &
   0.838 & 0.893 & 0.856 & 0.067 &
   0.809 & \underline{0.874} & \underline{0.805} & \textbf{0.067}
   \\

   \multicolumn{2}{c|}{$\text{GCoNet+}$ 
   } 
   & C9 
   &
   \underline{0.717} & \underline{0.798} & \underline{0.605} & 0.098 &
   0.853 & \underline{0.902} & 0.857 & 0.073 &
   \underline{0.819} & \textbf{0.877} & 0.796 & 0.075
   \\

   \rowcolor[RGB]{214,220,229}
   \multicolumn{2}{c|}{\textbf{$\text{CONDA}$}} 
   % & -
   & C9 
   &
   \textbf{0.730} & \textbf{0.801} & \textbf{0.622} & \underline{0.092} &
   \textbf{0.865} & \textbf{0.910} & \textbf{0.875} &
   \underline{0.059} &
   \textbf{0.825} & \textbf{0.877} & \textbf{0.810} & \underline{0.068}

   \\  

   \midrule

   \multicolumn{2}{c|}{$\text{CADC}$ 
   }
   & DC+C9 
   &
   0.680 & 0.744 & 0.549 & 0.133 &
   0.867 & 0.906 & 0.865 & 0.064 &
   0.815 & 0.854 & 0.778 & 0.088 
   \\   
   
   \multicolumn{2}{c|}{$\text{DMT}$ 
   } 
   & DC+C9 
   &
   0.725 & 0.800 & 0.619 & 0.108 &
   \underline{0.897} & \underline{0.936} & \underline{0.905} & \underline{0.045} &
   \underline{0.851} & \underline{0.895} & \underline{0.835} & \underline{0.063} 
   \\

   \multicolumn{2}{c|}{$\text{GCoNet+}$ 
   } 
   & DC+C9 
   &
   \underline{0.734} & \underline{0.808} & \underline{0.626} & \textbf{0.088} &
   0.876 & 0.920 & 0.880 & 0.057 &
   0.839 & 0.894 & 0.822 & 0.064
   \\

   \rowcolor[RGB]{214,220,229}
   \multicolumn{2}{c|}{\textbf{$\text{CONDA}$}} 
   & DC+C9 
   &
   \textbf{0.757} & \textbf{0.825} & \textbf{0.675} & \underline{0.092} &
   \textbf{0.904} & \textbf{0.940} & \textbf{0.912} & \textbf{0.042} &
   \textbf{0.857} & \textbf{0.899} & \textbf{0.844} & \textbf{0.060} 
   \\  

   \midrule
   
   \multicolumn{2}{c|}{$\text{UGEM}$ 
   } 
   & DC+CS 
   &
   0.726 & 0.808 & 0.599 & 0.096 &
   \underline{0.885} & \underline{0.935} & 0.882 & \underline{0.051} &
   \underline{0.853} & \textbf{0.911} & 0.829 & \underline{0.060}
   \\

   \multicolumn{2}{c|}{$\text{GCoNet+}$ 
   } 
   & DC+CS 
   &
   \underline{0.738} & \underline{0.814} & \underline{0.637} & \textbf{0.081} &
   0.881 & 0.926 & \underline{0.891} & 0.055 &
   0.843 & 0.901 & \underline{0.834} & 0.061
   \\

   \rowcolor[RGB]{214,220,229}
   \multicolumn{2}{c|}{\textbf{$\text{CONDA}$}} 
   % & - 
   & DC+CS 
   &
   \textbf{0.763} & \textbf{0.839} & \textbf{0.685} & \underline{0.089} &
   \textbf{0.900} & \textbf{0.938} & \textbf{0.908} & \textbf{0.045} &
   \textbf{0.862} & \underline{0.911} & \textbf{0.853} & \textbf{0.056}
   \\

   \bottomrule
   
  \end{tabular}
  \label{SOTATab}
  \vspace{-6mm}
\end{table}

\begin{figure*}[t]
	\scriptsize
	\renewcommand{\tabcolsep}{0.5pt} % adjust horizontal space
	\renewcommand{\arraystretch}{0.9} % adjust vertical space
	\centering
        \begin{tabular}{cccccccccc}
	    &
            \multicolumn{3}{c}{\cellcolor[RGB]{255,230,153}$\bold{Accordion}$}
            &
            \multicolumn{3}{c}{
            \cellcolor[RGB]{244,177,131}$\bold{rolling_pin}$}
	    &
            \multicolumn{3}{c}{
            \cellcolor[RGB]{169,209,142}$\bold{Towel}$}
            \vspace{0.5mm}
	    \\
            \arrayrulecolor{red}
            \rotatebox[origin=c]{90}{Image}
            &
		\makecell[c]{\includegraphics[width=0.105\linewidth,height=0.09\linewidth]{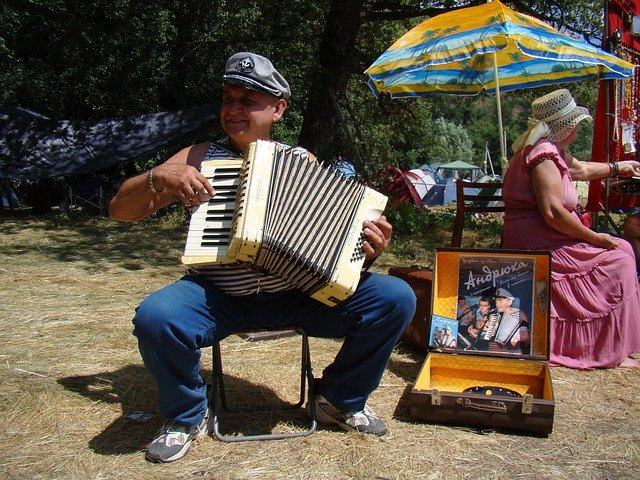}} 
		&
		\makecell[c]{\includegraphics[width=0.105\linewidth,height=0.09\linewidth]{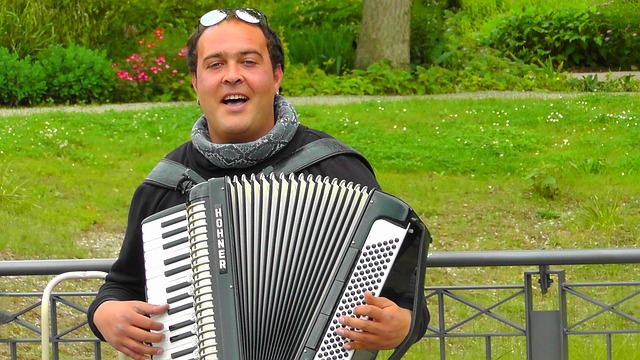}} 
		&
		\makecell[c]{\includegraphics[width=0.105\linewidth,height=0.09\linewidth]{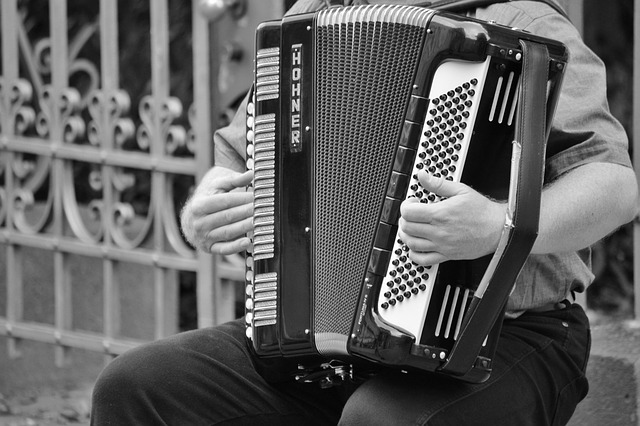}} 
		&
		\makecell[c]{\includegraphics[width=0.105\linewidth,height=0.09\linewidth]{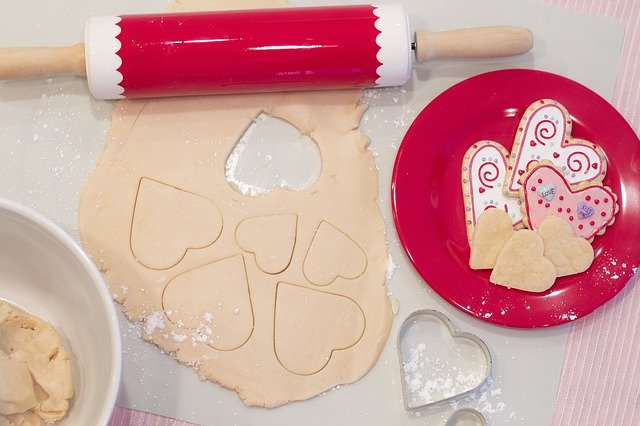}} 
		&
		\makecell[c]{\includegraphics[width=0.105\linewidth,height=0.09\linewidth]{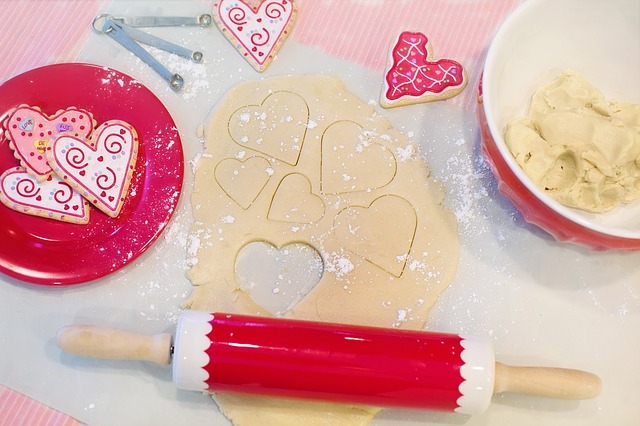}} 
		&
		\makecell[c]{\includegraphics[width=0.105\linewidth,height=0.09\linewidth]{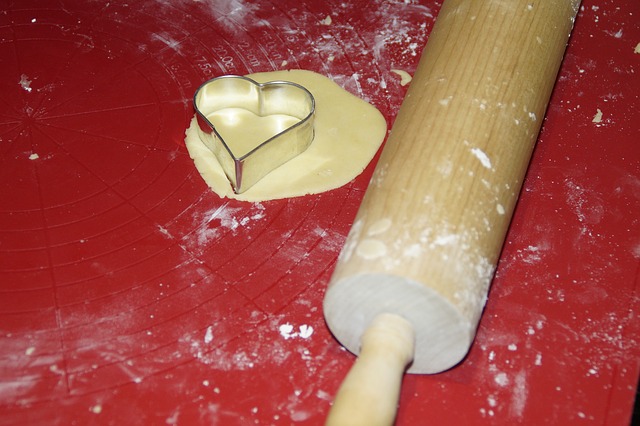}} 
		&
		\makecell[c]{\includegraphics[width=0.105\linewidth,height=0.09\linewidth]{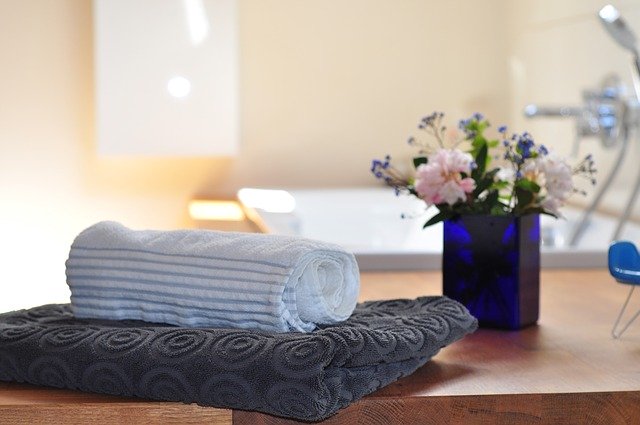}} 
		&
		\makecell[c]{\includegraphics[width=0.105\linewidth,height=0.09\linewidth]{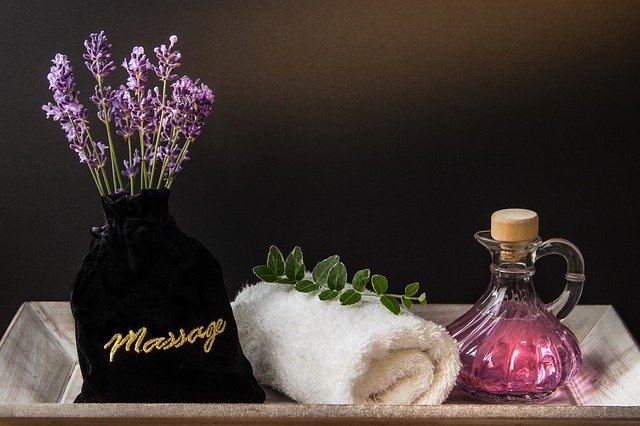}} 
		&
		\makecell[c]{\includegraphics[width=0.105\linewidth,height=0.09\linewidth]{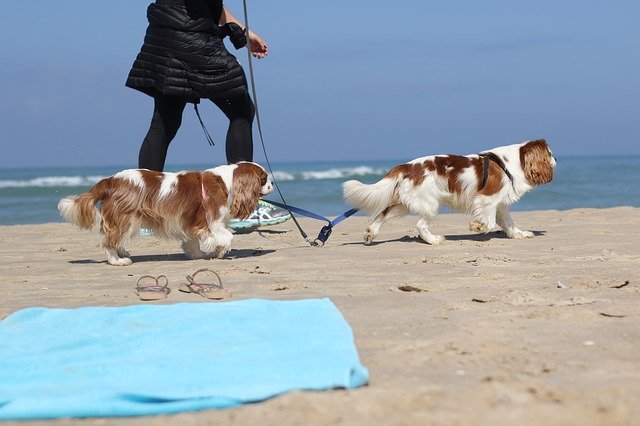}}

        \vspace{-0.5mm}
		\\
 
            \rotatebox[origin=c]{90}{GT}
            &
		\makecell[c]{\includegraphics[width=0.105\linewidth,height=0.09\linewidth]{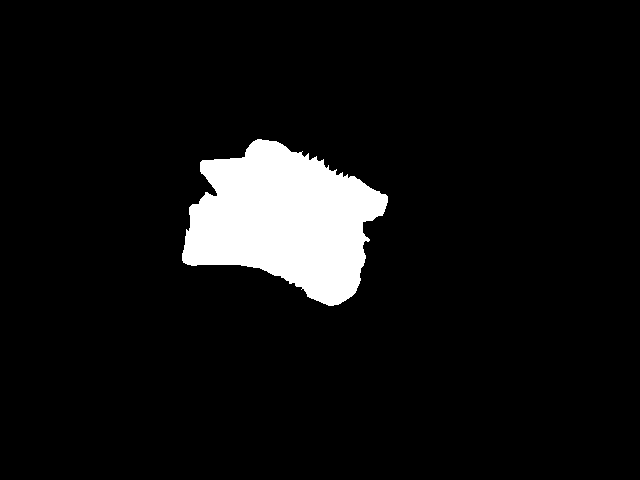}} 
		&
		\makecell[c]{\includegraphics[width=0.105\linewidth,height=0.09\linewidth]{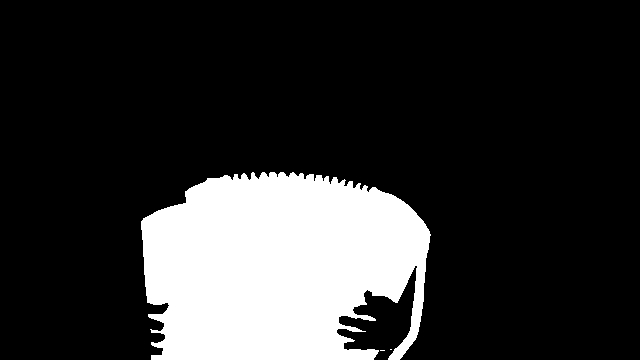}} 
		&
		\makecell[c]{\includegraphics[width=0.105\linewidth,height=0.09\linewidth]{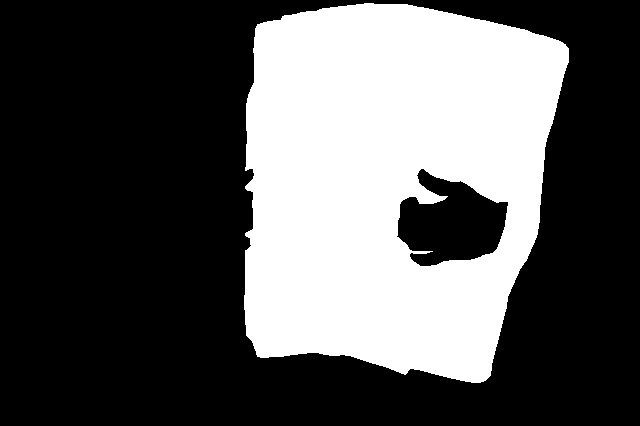}} 
		&
		\makecell[c]{\includegraphics[width=0.105\linewidth,height=0.09\linewidth]{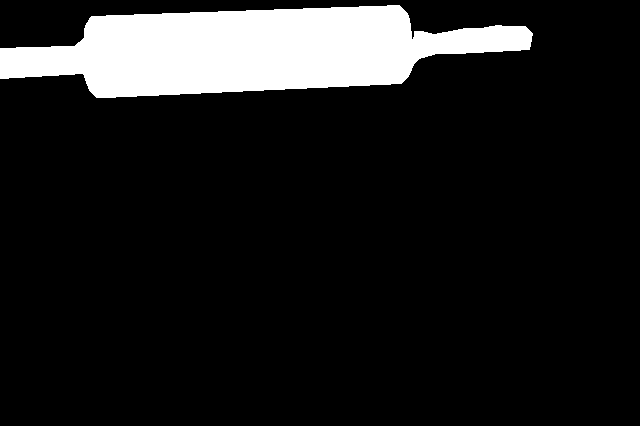}} 
		&
		\makecell[c]{\includegraphics[width=0.105\linewidth,height=0.09\linewidth]{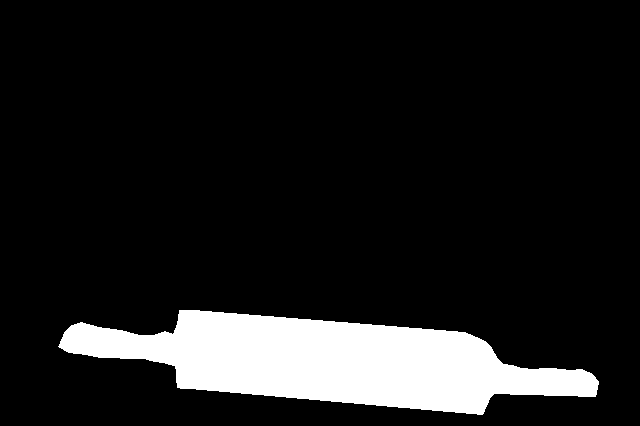}} 
		&
		\makecell[c]{\includegraphics[width=0.105\linewidth,height=0.09\linewidth]{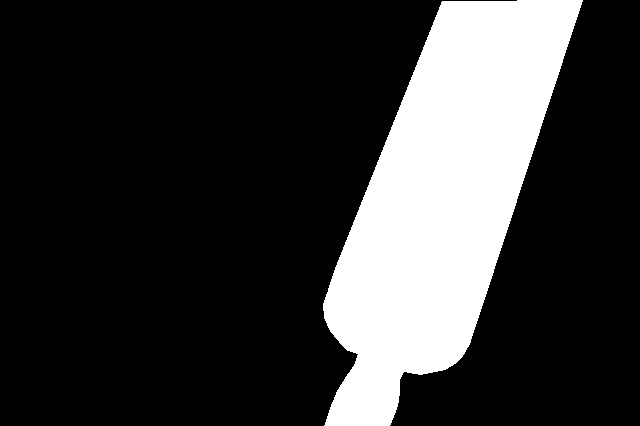}} 
		&
		\makecell[c]{\includegraphics[width=0.105\linewidth,height=0.09\linewidth]{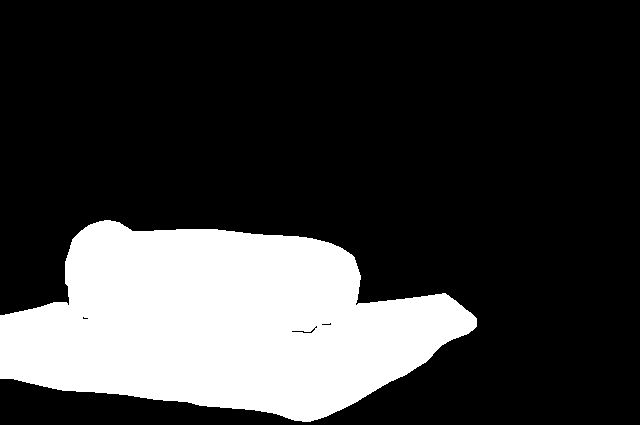}} 
		&
		\makecell[c]{\includegraphics[width=0.105\linewidth,height=0.09\linewidth]{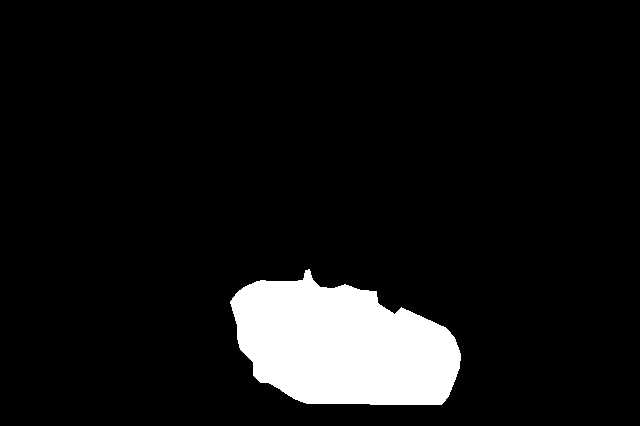}} 
		&
		\makecell[c]{\includegraphics[width=0.105\linewidth,height=0.09\linewidth]{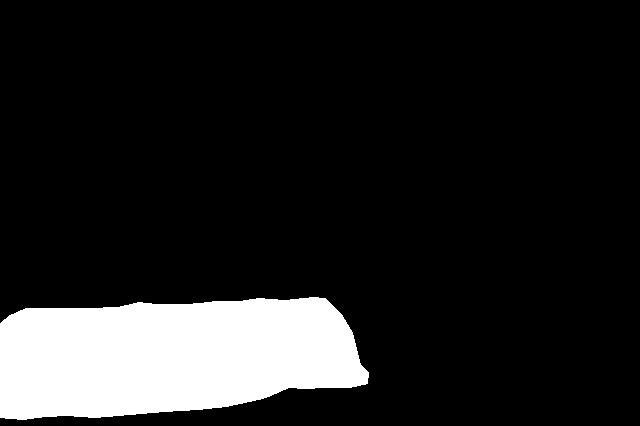}}
        \vspace{-0.5mm}
		\\

            \rotatebox[origin=c]{90}{$\red{Ours}$}
            &
		\makecell[c]{\includegraphics[width=0.105\linewidth,height=0.09\linewidth]{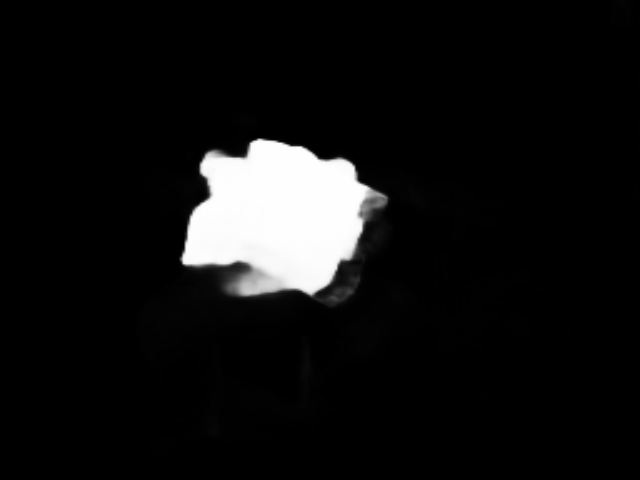}} 
		&
		\makecell[c]{\includegraphics[width=0.105\linewidth,height=0.09\linewidth]{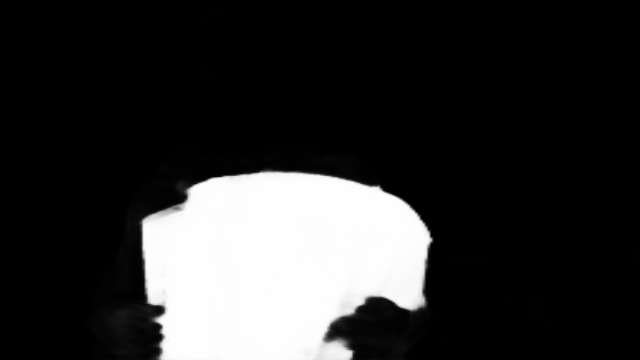}} 
		&
		\makecell[c]{\includegraphics[width=0.105\linewidth,height=0.09\linewidth]{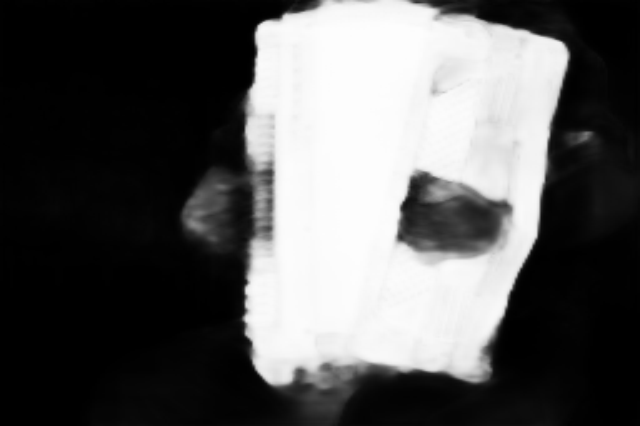}} 
		&
		\makecell[c]{\includegraphics[width=0.105\linewidth,height=0.09\linewidth]{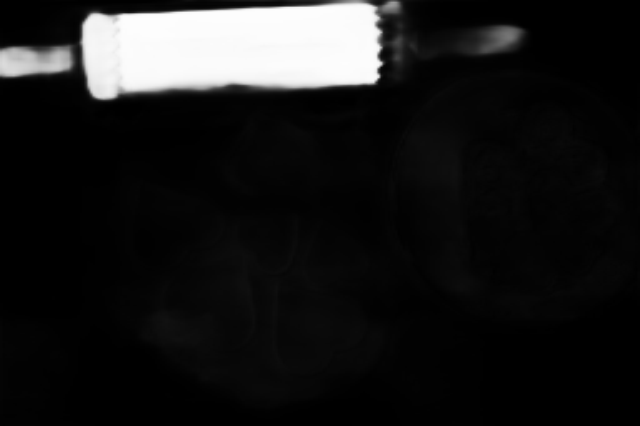}} 
		&
		\makecell[c]{\includegraphics[width=0.105\linewidth,height=0.09\linewidth]{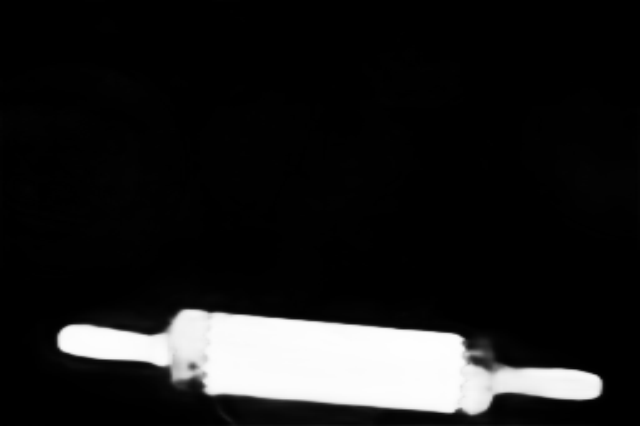}} 
		&
		\makecell[c]{\includegraphics[width=0.105\linewidth,height=0.09\linewidth]{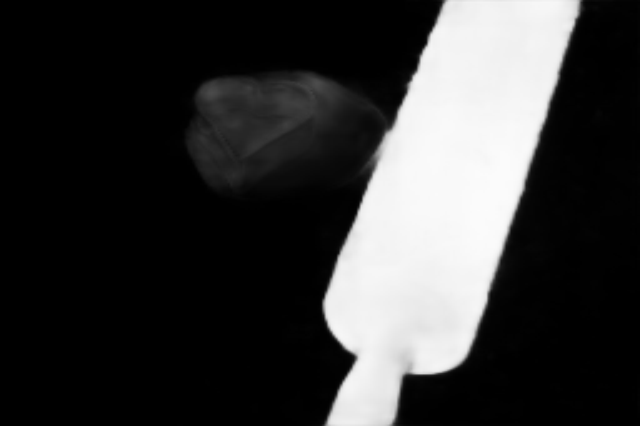}} 
		&
		\makecell[c]{\includegraphics[width=0.105\linewidth,height=0.09\linewidth]{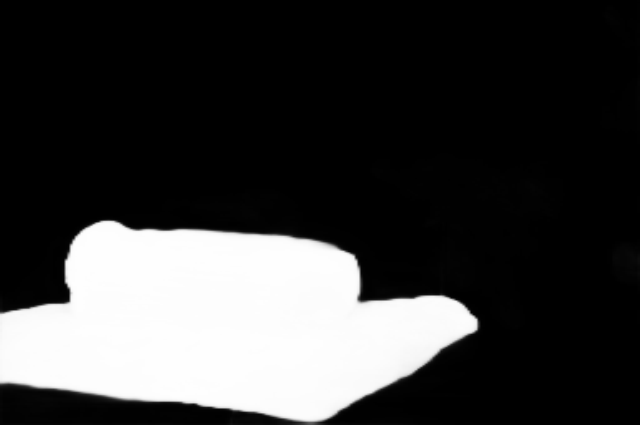}} 
		&
		\makecell[c]{\includegraphics[width=0.105\linewidth,height=0.09\linewidth]{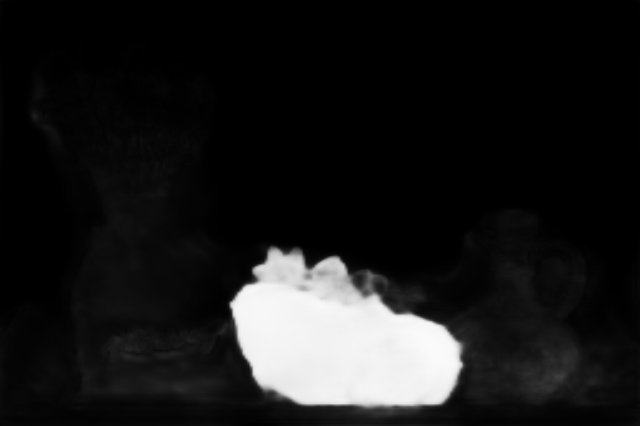}} 
		&
		\makecell[c]{\includegraphics[width=0.105\linewidth,height=0.09\linewidth]{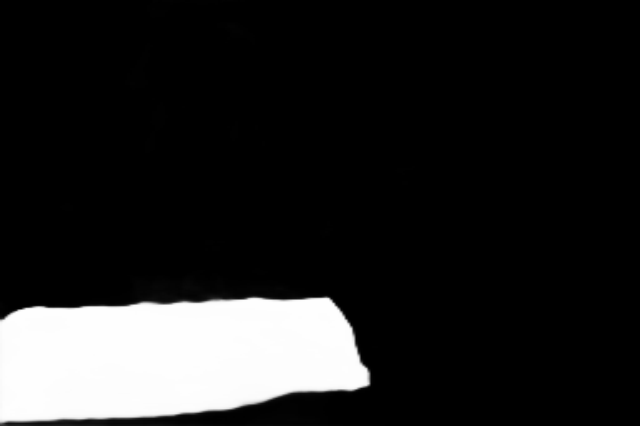}}         \vspace{-0.5mm}
		\\

            \rotatebox[origin=c]{90}{GCoNet+}
            &
		\makecell[c]{\includegraphics[width=0.105\linewidth,height=0.09\linewidth]{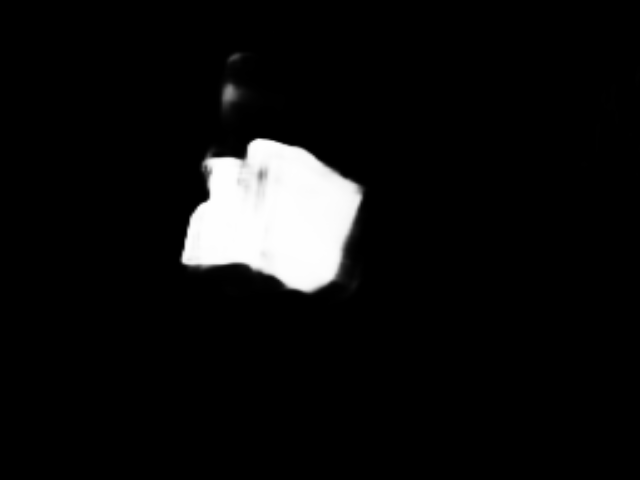}} 
		&
		\makecell[c]{\includegraphics[width=0.105\linewidth,height=0.09\linewidth]{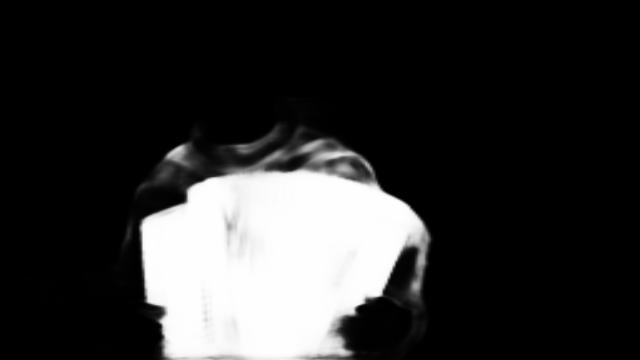}} 
		&
		\makecell[c]{\includegraphics[width=0.105\linewidth,height=0.09\linewidth]{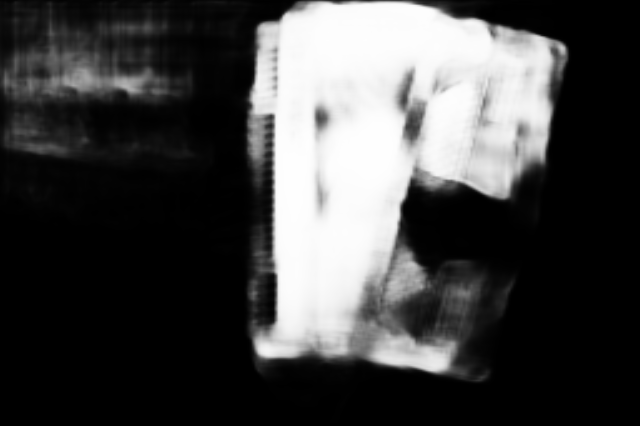}} 
		&
		% \makecell[c]{\includegraphics[width=0.105\linewidth,height=0.09\linewidth]{figures/sota_maps/CoCA/Accordion/4664279/10_gconet2_0.8856.png}} 
		% &
		\makecell[c]{\includegraphics[width=0.105\linewidth,height=0.09\linewidth]{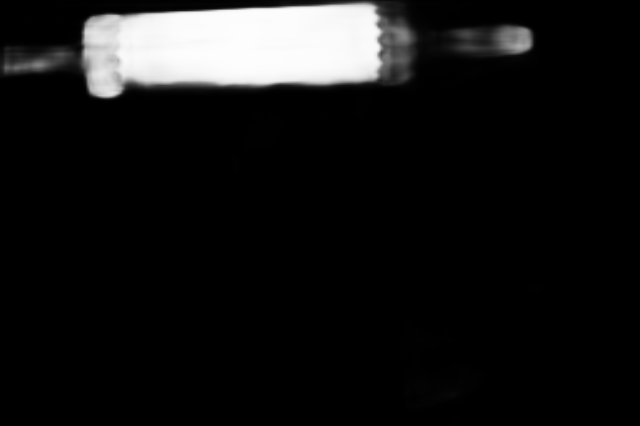}} 
		&
		\makecell[c]{\includegraphics[width=0.105\linewidth,height=0.09\linewidth]{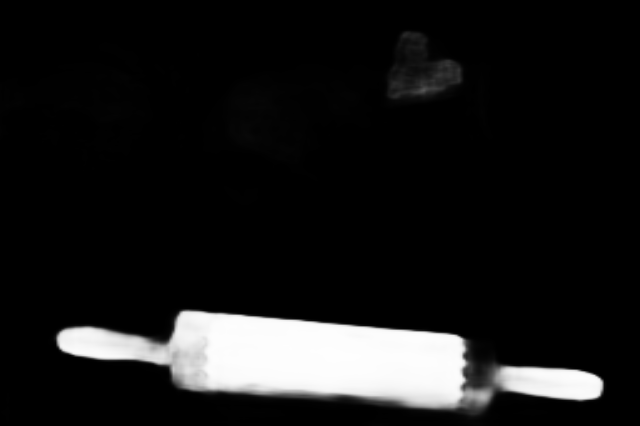}} 
		&
		\makecell[c]{\includegraphics[width=0.105\linewidth,height=0.09\linewidth]{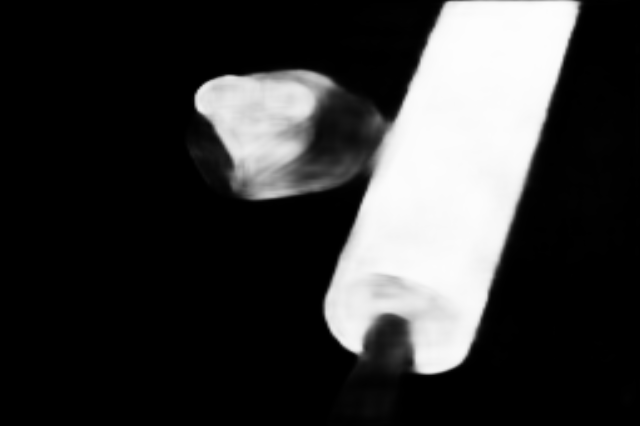}} 
		&
		\makecell[c]{\includegraphics[width=0.105\linewidth,height=0.09\linewidth]{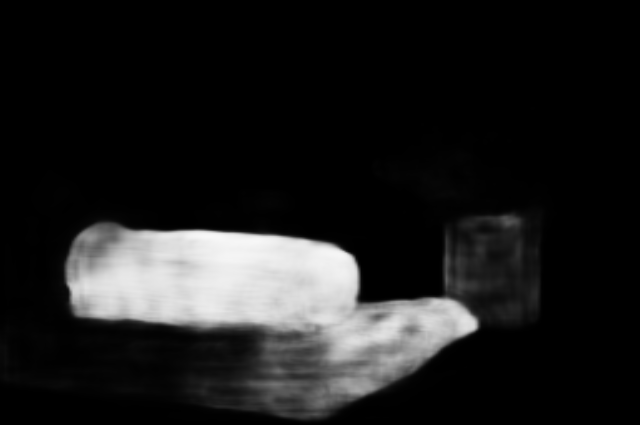}} 
		&
		\makecell[c]{\includegraphics[width=0.105\linewidth,height=0.09\linewidth]{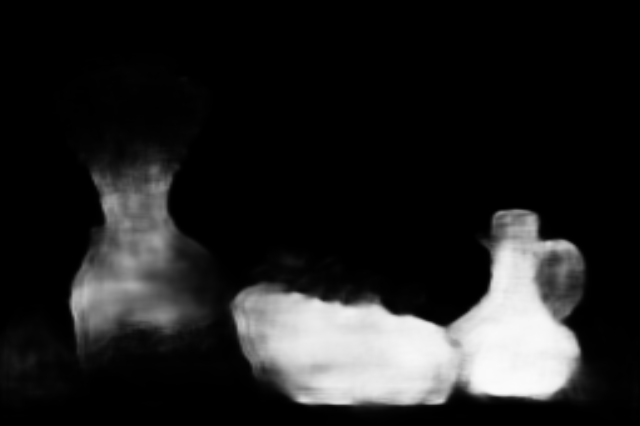}} 
		&
		\makecell[c]{\includegraphics[width=0.105\linewidth,height=0.09\linewidth]{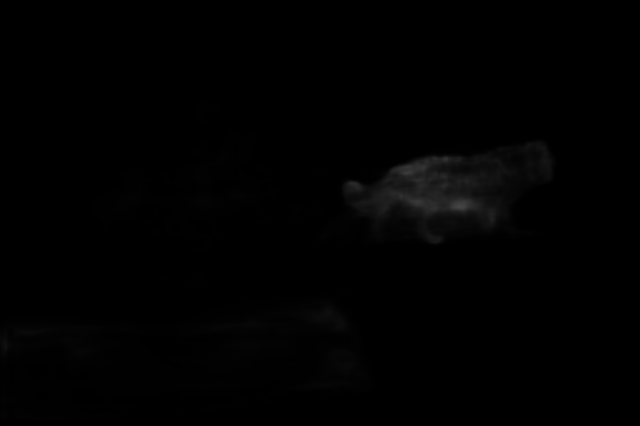}}
		% \makecell[c]{\includegraphics[width=0.105\linewidth,height=0.09\linewidth]{figures/sota_maps/CoSOD3k/ant/ILSVRC2012_val_00011669/10_gconet2_0.8242.png}} 
		% &
		% \makecell[c]{\includegraphics[width=0.105\linewidth,height=0.09\linewidth]{figures/sota_maps/CoSOD3k/ant/ILSVRC2012_val_00021864/10_gconet2_0.8092.png}} 
		% &
		% \makecell[c]{\includegraphics[width=0.105\linewidth,height=0.09\linewidth]{figures/sota_maps/CoSOD3k/ant/ILSVRC2012_val_00029818/10_gconet2_0.9031.png}} 
		% &
		% \makecell[c]{\includegraphics[width=0.105\linewidth,height=0.09\linewidth]{figures/sota_maps/CoSOD3k/ant/ILSVRC2012_val_00035087/10_gconet2_0.8686.png}} 
            \vspace{-0.5mm}
		\\

            \rotatebox[origin=c]{90}{UGEM}
            &
		\makecell[c]{\includegraphics[width=0.105\linewidth,height=0.09\linewidth]{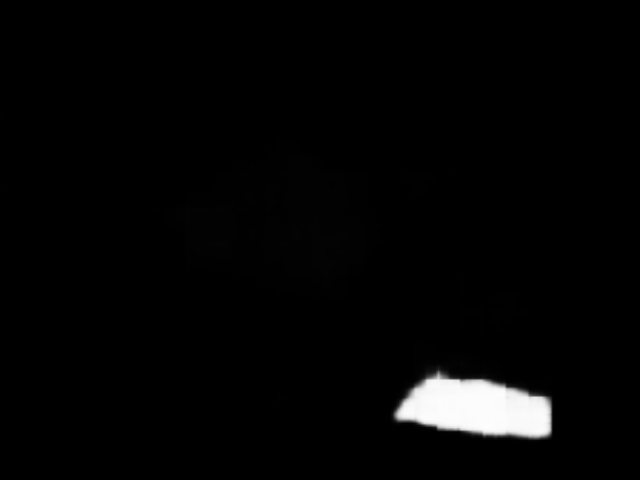}} 
		&
		\makecell[c]{\includegraphics[width=0.105\linewidth,height=0.09\linewidth]{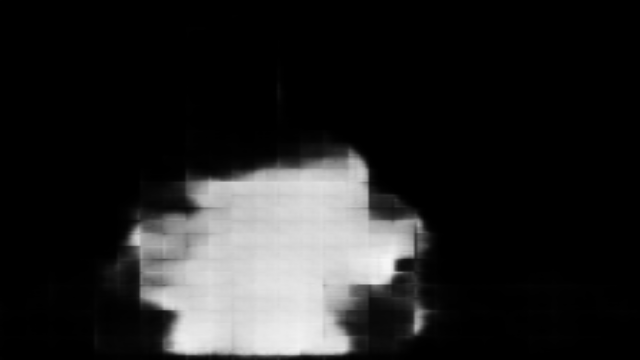}} 
		&
		\makecell[c]{\includegraphics[width=0.105\linewidth,height=0.09\linewidth]{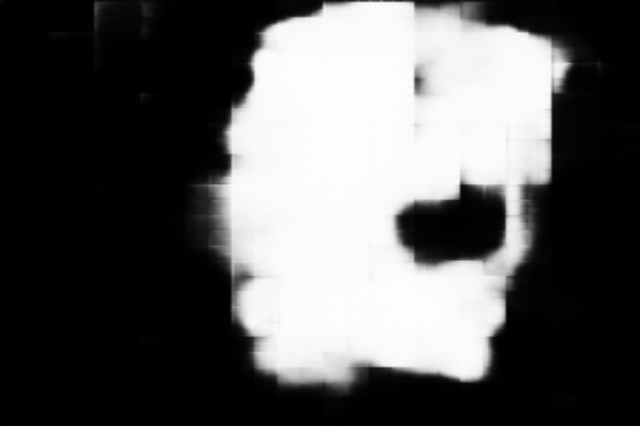}} 
		&
		\makecell[c]{\includegraphics[width=0.105\linewidth,height=0.09\linewidth]{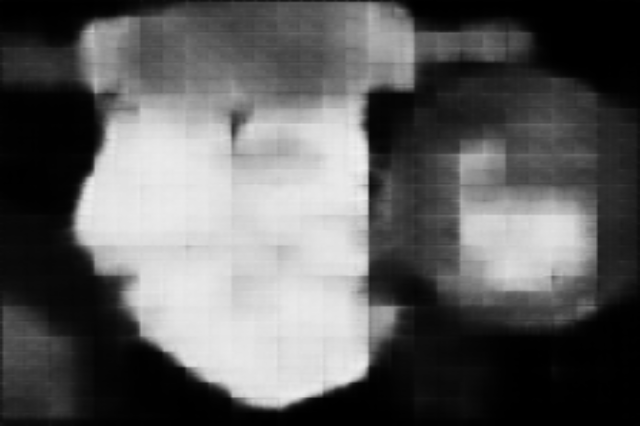}} 
		&
		\makecell[c]{\includegraphics[width=0.105\linewidth,height=0.09\linewidth]{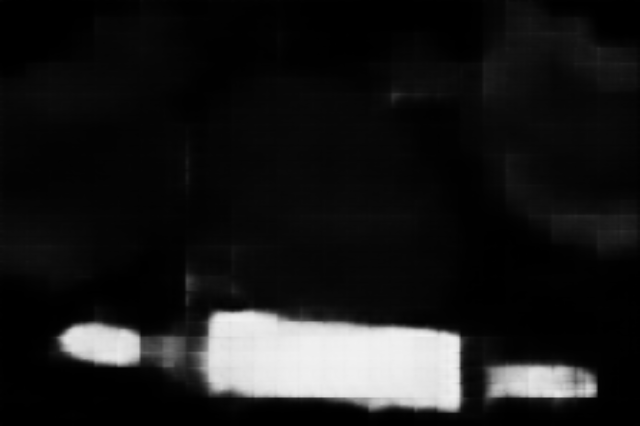}} 
		&
		\makecell[c]{\includegraphics[width=0.105\linewidth,height=0.09\linewidth]{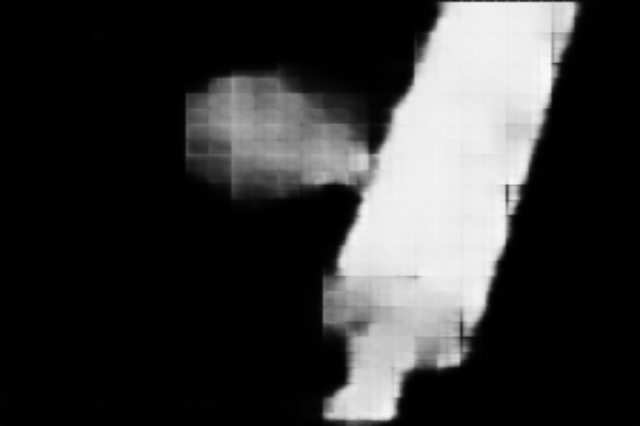}} 
		&
		\makecell[c]{\includegraphics[width=0.105\linewidth,height=0.09\linewidth]{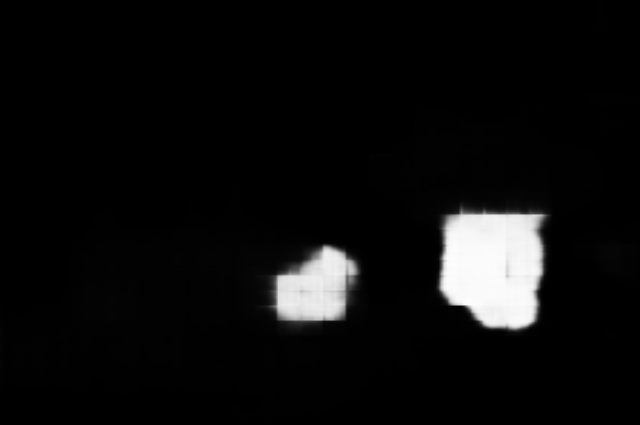}} 
		&
		\makecell[c]{\includegraphics[width=0.105\linewidth,height=0.09\linewidth]{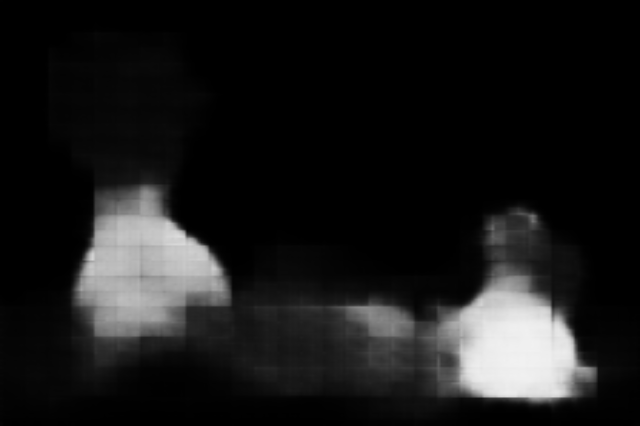}} 
		&
		\makecell[c]{\includegraphics[width=0.105\linewidth,height=0.09\linewidth]{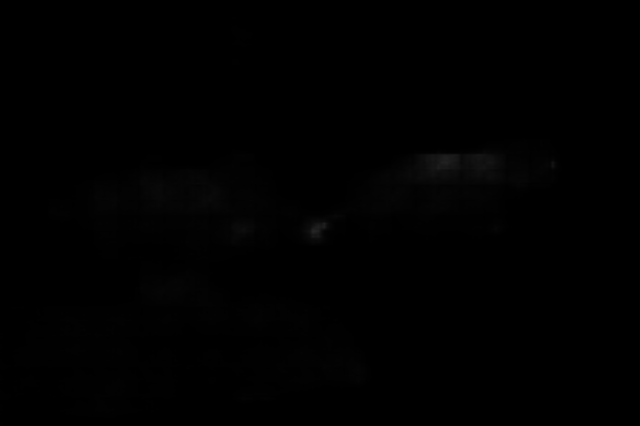}}
		% \makecell[c]{\includegraphics[width=0.105\linewidth,height=0.09\linewidth]{figures/sota_maps/CoSOD3k/ant/ILSVRC2012_val_00011669/9_ugem_0.8392.png}} 
		% &
		% \makecell[c]{\includegraphics[width=0.105\linewidth,height=0.09\linewidth]{figures/sota_maps/CoSOD3k/ant/ILSVRC2012_val_00021864/9_ugem_0.8316.png}} 
		% &
		% \makecell[c]{\includegraphics[width=0.105\linewidth,height=0.09\linewidth]{figures/sota_maps/CoSOD3k/ant/ILSVRC2012_val_00029818/9_ugem_0.8368.png}} 
		% &
		% \makecell[c]{\includegraphics[width=0.105\linewidth,height=0.09\linewidth]{figures/sota_maps/CoSOD3k/ant/ILSVRC2012_val_00035087/9_ugem_0.8428.png}} 
            \vspace{-0.5mm}
		\\

            \rotatebox[origin=c]{90}{DMT}
            &
		\makecell[c]{\includegraphics[width=0.105\linewidth,height=0.09\linewidth]{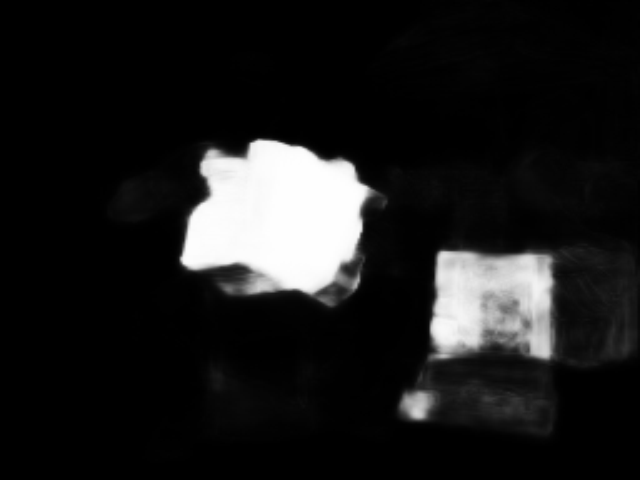}} 
		&
		\makecell[c]{\includegraphics[width=0.105\linewidth,height=0.09\linewidth]{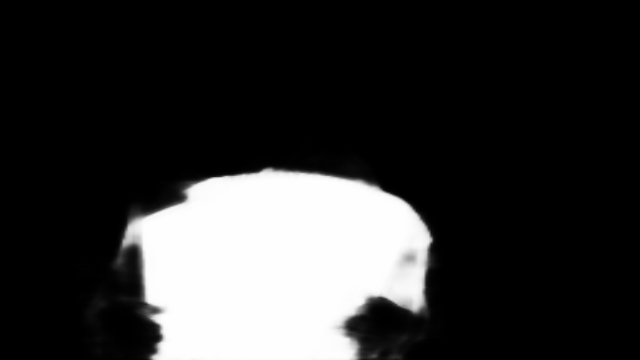}} 
		&
		\makecell[c]{\includegraphics[width=0.105\linewidth,height=0.09\linewidth]{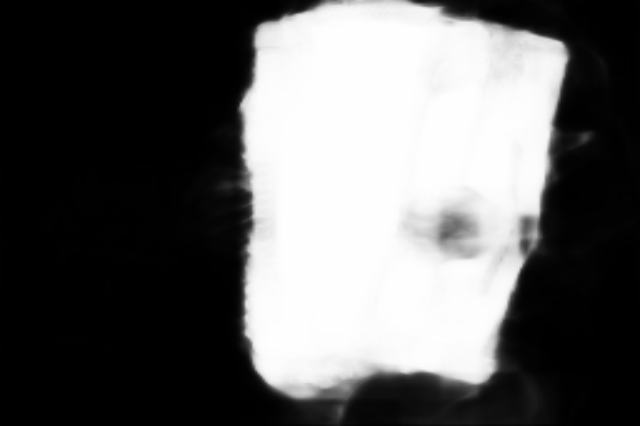}} 
		&
		\makecell[c]{\includegraphics[width=0.105\linewidth,height=0.09\linewidth]{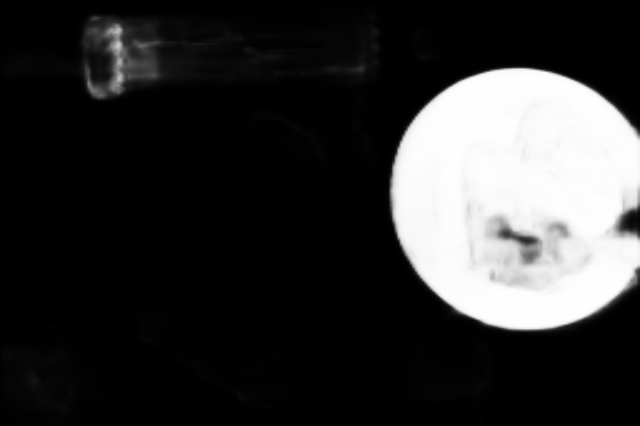}} 
		&
		\makecell[c]{\includegraphics[width=0.105\linewidth,height=0.09\linewidth]{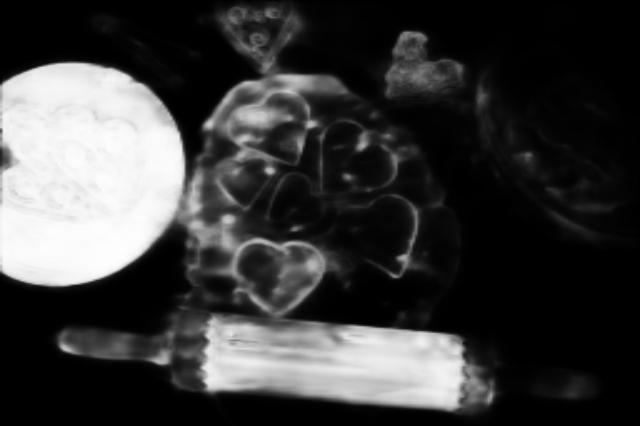}} 
		&
		\makecell[c]{\includegraphics[width=0.105\linewidth,height=0.09\linewidth]{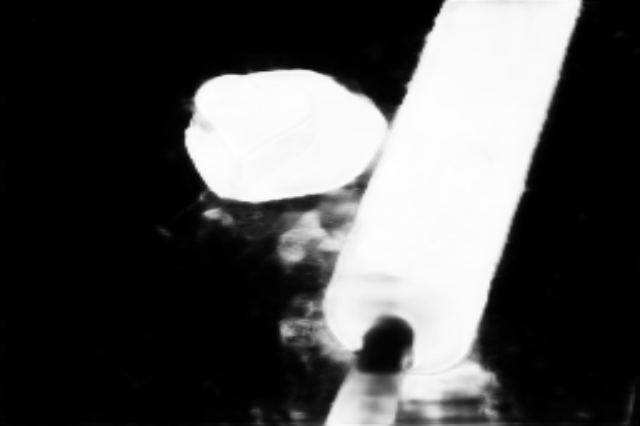}} 
		&
		\makecell[c]{\includegraphics[width=0.105\linewidth,height=0.09\linewidth]{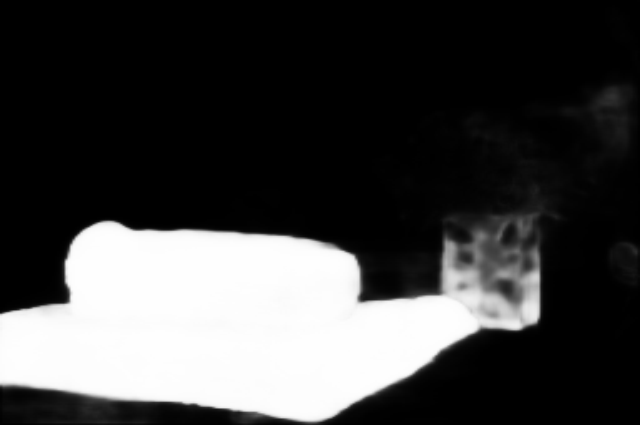}} 
		&
		\makecell[c]{\includegraphics[width=0.105\linewidth,height=0.09\linewidth]{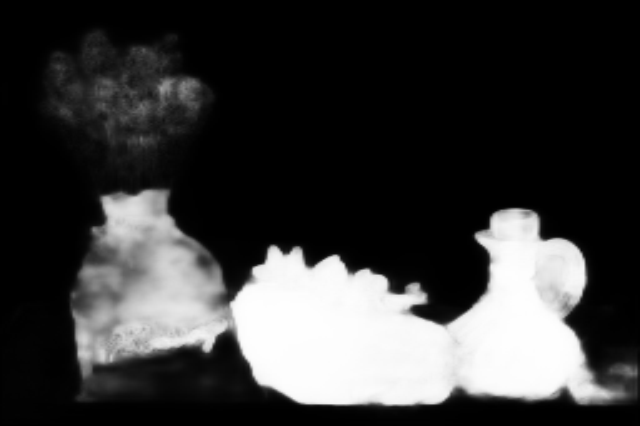}} 
		&
		\makecell[c]{\includegraphics[width=0.105\linewidth,height=0.09\linewidth]{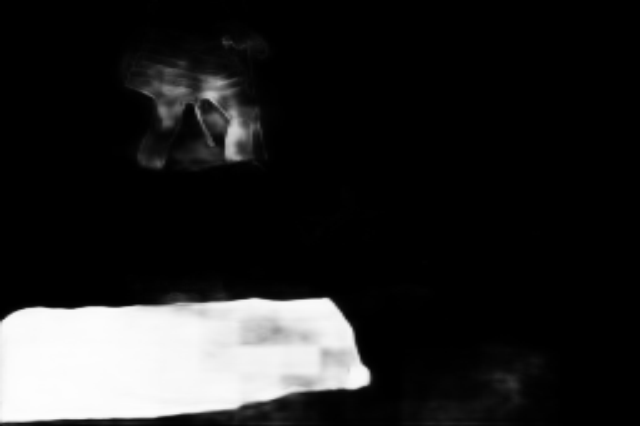}}
		% \makecell[c]{\includegraphics[width=0.105\linewidth,height=0.09\linewidth]{figures/sota_maps/CoSOD3k/ant/ILSVRC2012_val_00011669/8_dmt_0.8540.png}} 
		% &
		% \makecell[c]{\includegraphics[width=0.105\linewidth,height=0.09\linewidth]{figures/sota_maps/CoSOD3k/ant/ILSVRC2012_val_00021864/8_dmt_0.8496.png}} 
		% &
		% \makecell[c]{\includegraphics[width=0.105\linewidth,height=0.09\linewidth]{figures/sota_maps/CoSOD3k/ant/ILSVRC2012_val_00029818/8_dmt_0.9015.png}} 
		% &
		% \makecell[c]{\includegraphics[width=0.105\linewidth,height=0.09\linewidth]{figures/sota_maps/CoSOD3k/ant/ILSVRC2012_val_00035087/8_dmt_0.8998.png}} 
            \vspace{-0.5mm}
		\\

            \vspace{-0.5mm}
		\\

		\end{tabular}
            \vspace{-4mm}
		\caption{
		\textbf{Qualitative comparisons of our model with other SOTA methods.}
		}
		\label{SOTAFig}
		\vspace{-6mm}
\end{figure*}

\subsection{Comparison with State-of-the-Art Methods}
We compare our model with eight recent SOTA methods, \ie GICD \cite{zhang2020gicd}, ICNet \cite{jin2020icnet}, GCoNet \cite{fan2021GCoNet}, CADC \cite{zhang2021summarize}, DCFM \cite{yu2022democracy}, DMT \cite{li2023discriminative}, UGEM \cite{wu2023co}, and GCoNet+ \cite{zheng2023gconet+}. 
We directly utilize their officially released saliency maps for comparison. To ensure fairness, we trained our model with different combinations of three training datasets, following \cite{zheng2023gconet+}, to align with other compared methods.
We denote three training datasets, \ie DUTS class \cite{zhang2020gicd}, COCO9k \cite{lin2014microsoft}, and COCO-SEG \cite{wang2019robust}, as DC, C9, and CS, respectively, for convenience. Our training sets include DC, C9, DC+C9, DC+CS. 
As shown in Table~\ref{SOTATab}, 
we can observe that our model achieves the best performance with each training set in most benchmark datasets. What is even more exciting is that we achieve excellent results in the most challenging CoCA dataset, surpassing the second-best models by large margins, \eg 2.5\% $S_m$, 2.5\% $E_\xi$, and 4.8\% $\text{F}_\beta$ with the DC+CS training set.  

We also present visual comparisons in Figure~\ref{SOTAFig}. Our model can accurately detect co-salient objects in complex scenarios, such as irregularly shaped accordions accompanied by extraneous objects (people). However, other models easily fail to accurately segment co-salient objects.

\section{Conclusion}
This paper proposes a deep association learning strategy for CoSOD that directly
embeds hyperassociations into deep association features. Correspondence
estimation is also introduced to condense hyperassociations, enabling the exploration
of pixel-level correspondences for CoSOD. We also utilize an object-aware
cycle consistency loss to further refine correspondence estimations. Extensive
experiments have verified the effectiveness of our method.

\Paragraph{Acknowledgments.} This work was supported in part by Key-Area Research and Development Program of
Guangdong Province under
Grant 2021B0101200001, the National Natural Science Foundation of China under Grant 62136007, 62036011, U20B2065, 6202781, 62036005, the Key R\&D Program of Shaanxi Province under Grant 2021ZDLGY01-08, and 
the MBZUAI-WIS Joint Program for AI Research under Grants WIS P008 and P009.

\bibliographystyle{splncs04}
% \bibliography{main}

\end{document}